\documentclass[journal]{IEEEtran}

\usepackage{stmaryrd}
\usepackage{bm}
\usepackage{amssymb}
\usepackage{booktabs}
\usepackage{graphicx}
\usepackage{float}
\usepackage{graphicx}
\usepackage{stfloats}
\usepackage{amsmath}
\usepackage{epstopdf}
\usepackage{algorithm,algorithmic}
\usepackage{multirow}
\usepackage{bbm}
\usepackage{framed}
\usepackage[colorlinks,linkcolor=red,anchorcolor=green,citecolor=green]{hyperref}
\usepackage[numbers,sort&compress]{natbib}
\usepackage{amsthm}

\newcommand{\ie}{{\em i.e.}}

\newtheorem{definition}{Definition}
\newtheorem{lemma}{Lemma}
\newtheorem{corollary}{Corollary}
\newtheorem{theorem}{Theorem}
\ifCLASSINFOpdf

\else

\fi

\begin{document}

\title{A Benchmark for Sparse Coding: When \\Group Sparsity Meets Rank Minimization}

\author{Zhiyuan~Zha,~\IEEEmembership{Member,~IEEE,} Xin~Yuan,~\IEEEmembership{Senior Member,~IEEE,} Bihan~Wen,~\IEEEmembership{Member,~IEEE,} Jiantao~Zhou,~\IEEEmembership{Member,~IEEE}, Jiachao~Zhang,~\IEEEmembership{Member,~IEEE} and Ce~Zhu,~\IEEEmembership{Fellow,~IEEE}

\IEEEcompsocitemizethanks{\IEEEcompsocthanksitem This work was supported by the NSFC (61571102), the applied research programs of science and technology., Sichuan Province (No. 2018JY0035), the Ministry of Education, Republic of Singapore, under the Start-up Grant and the Macau Science and Technology Development Fund, Macau SAR (File no. SKL-IOTSC-2018-2020, 077/2018/A2, 022/2017/A1).
\IEEEcompsocthanksitem Z. Zha and C. Zhu are with the School of Information and Communication Engineering,
University of Electronic Science and Technology of China, Chengdu, 611731, China.  E-mail: zhazhiyuan.mmd@gmail.com, eczhu@uestc.edu.cn. \emph{Corresponding Author: Ce Zhu.}
\IEEEcompsocthanksitem X. Yuan is with Nokia Bell Labs, 600 Mountain Avenue, Murray Hill, NJ 07974, USA. E-mail: xyuan@bell-labs.com.
\IEEEcompsocthanksitem B. Wen is with School of Electrical and Electronic Engineering, Nanyang Technological University, Singapore 639798. E-mail: bihan.wen@ntu.edu.sg.
\IEEEcompsocthanksitem J. Zhou is with Department of Computer and Information Science,  University of Macau, Macau 999078, China. E-mail: jtzhou@umac.mo.
\IEEEcompsocthanksitem J. Zhang is with Kangni Mechanical and Electrical Institute, Nanjing Institute of Technology, Nanjing 211167, China. E-mail: zhangjc07@foxmail.com.
}
}
\markboth{IEEE Transaction on Image Processing,~2020}%
{Shell \MakeLowercase{\textit{et al.}}: Bare Demo of IEEEtran.cls for IEEE Journals}

\maketitle

\begin{abstract}
Sparse coding has achieved a great success in various image processing tasks. However, a benchmark to measure the sparsity of image patch/group is missing since sparse coding is essentially an NP-hard problem. This work attempts to fill the gap from the perspective of rank minimization. We firstly design an adaptive dictionary to bridge the gap between group-based sparse coding (GSC) and rank minimization. Then, we show that under the designed dictionary, GSC and the rank minimization problems are equivalent, and therefore the sparse coefficients of each patch group can be measured by estimating the singular values of each patch group. We thus earn a benchmark to measure the sparsity of each patch group because the singular values of the original image patch groups can be easily computed by the singular value decomposition (SVD). This benchmark can be used to evaluate performance of any kind of norm minimization methods in sparse coding through analyzing their corresponding rank minimization counterparts.  Towards this end, we exploit four well-known rank minimization methods to study the sparsity of each patch group and the weighted Schatten $p$-norm minimization (WSNM) is found to be the closest one to the real singular values of each patch group. Inspired by the aforementioned equivalence regime of rank minimization and GSC, WSNM can be translated into a non-convex weighted $\ell_p$-norm minimization problem in GSC. By using the earned benchmark in sparse coding, the weighted $\ell_p$-norm minimization is expected to obtain  better performance than the three other norm minimization methods, \ie, $\ell_1$-norm, $\ell_p$-norm and weighted $\ell_1$-norm.
To verify the feasibility of the proposed benchmark, we compare the weighted $\ell_p$-norm minimization against the three aforementioned norm minimization methods in sparse coding. Experimental results on image restoration applications, namely image inpainting and image compressive sensing recovery, demonstrate that the proposed scheme is feasible and outperforms many state-of-the-art methods.
\end{abstract}

\begin{IEEEkeywords}
Sparse coding, GSC, rank minimization, adaptive dictionary, weighted $\ell_p$-norm minimization, image restoration, compressive sensing, nuclear norm.
\end{IEEEkeywords}

\IEEEpeerreviewmaketitle

\section{Introduction}
\label{sec:1}

\IEEEPARstart{T}{raditional} patch-based sparse coding (PSC) has been widely used in various image processing tasks and has achieved excellent results \cite{1,2,3,4}. It assumes that each patch of an image can be precisely modeled by a sparse linear combination of some fixed and trainable basis elements, which are called atoms and these atoms compose a dictionary. As such, one key issue in sparse coding based scheme is to train a dictionary, with popular techniques including K-SVD \cite{1}, ODL \cite{3} and SDL \cite{4}. 
However, there exists two main issues in the PSC model. $i)$ Since dictionary learning is a large-scale and highly non-convex problem, it is computationally expensive. $ii)$ The PSC model usually assumes the independence of image patches, which does not take account of the correlation within similar patches \cite{6,7,8,5}.

To overcome the aforementioned limitations, instead of using a single patch as the basic unit in sparse coding, recent advances of group-based sparse coding (GSC) consider similar {\em patch group} as the basic unit and have demonstrated great potentials in various image processing tasks \cite{6,7,8,5,10,11,12,14,72}. The GSC is a powerful mechanism to integrate intrinsic local sparsity and nonlocal self-similarity (NSS) of images \cite{6}. Specifically, taking an (vectorized) image $\textbf{\emph{x}}\in{\mathbb R}^{N}$ as an example, it is divided into $n$ overlapped patches of size $\sqrt{d}\times \sqrt{d}$, and each patch is denoted by a vector ${\textbf{\emph{x}}}_i\in\mathbb{R}^{d}, i=1,2,\dots,n$. Then for each patch ${\textbf{\emph{x}}}_i$,  $m$ similar matched patches are selected from a searching window with $C \times C$ pixels to form a set ${\textbf{\emph{S}}}_i$.  It is noticed that K-Nearest Neighbour (KNN) algorithm \cite{15} is usually utilized to search similar matched patches. Following this, all patches in ${\textbf{\emph{S}}}_i$ are stacked into a matrix ${\textbf{\emph{X}}}_i\in\mathbb{R}^{d\times m}$, \ie, ${\textbf{\emph{X}}}_i=\{{\textbf{\emph{x}}}_{i,1}, {\textbf{\emph{x}}}_{i,2},...,{\textbf{\emph{x}}}_{i,m}\}$.
This matrix ${\textbf{\emph{X}}}_i$ consisting of patches with similar structures is thereby called a patch group, where $\{{\textbf{\emph{x}}}_{i,j}\}_{j=1}^m$ denotes the $j$-th patch in the $i$-th patch group.  Similar to PSC \cite{1,2}, given a dictionary ${\textbf{\emph{D}}}_{i}$, each patch group ${\textbf{\emph{X}}}_i$ can be sparsely represented by solving the following $\ell_0$-norm minimization problem,
\begin{equation}
\hat{\textbf{\emph{A}}}_i=\arg\min_{{\textbf{\emph{A}}}_i} \left(\frac{1}{2}\left\|{\textbf{\emph{X}}}_i-{\textbf{\emph{D}}}_i{\textbf{\emph{A}}}_i\right\|_F^2+\lambda\left\|{\textbf{\emph{A}}}_i\right\|_0\right),
\label{eq:2}
\end{equation} 
where ${\textbf{\emph{A}}}_i$ represents the group sparse coefficient of each patch group ${\textbf{\emph{X}}}_i$ and $\lambda$ is a regularization parameter. $\left\|~\right\|_F$ denotes the Frobenius norm, and $\left\|~\right\|_0$ signifies the $\ell_0$-norm
, \ie, counting the nonzero entries of each column in ${\textbf{\emph{A}}}_i$. Then the entire image ${\textbf{\emph{x}}}$ can be sparsely represented by a set of {\em group sparse codes} $\{{\textbf{\emph{A}}}_i\}_{i=1}^n$.

It is well-known that norm minimization methods are often used to evaluate the sparsity of a signal/image \cite{16,17,18,19,20,21,22,23}.  However, since $\ell_0$-norm minimization problem is a difficult combinatorial optimization, solving Eq.~\eqref{eq:2} is NP-hard. For this reason, it is often replaced by the $\ell_1$-norm minimization \cite{16} to make the optimization problem tractable. 
Meanwhile, advanced norm minimization methods, like the weighted $\ell_1$-norm minimization \cite{17}, the $\ell_{1/2}$-norm minimization \cite{18} and the weighted $\ell_{2, 1}$-norm minimization \cite{19}, are also proposed to solve the $\ell_0$-norm minimization problem. Unfortunately, for some practical problems, such as image inverse problems \cite{6,7,8,10}, no matter which method is used, it is just an {\em estimate} to the $\ell_0$-norm minimization and cannot obtain the real sparse solution. During the past decades, there are still existing many arguments about which one is the best (in the sparsity approximation perspective) of these norm minimization methods \cite{17,18,19,20,21,22,23}. 
More importantly, although we possess the original signal/image $\textbf{\emph{x}}$, it is almost impossible to obtain the real solution of each group sparse coefficient ${\textbf{\emph{A}}}_i$ in Eq.~\eqref{eq:2}, due to the NP-hard essence of the  $\ell_0$-norm minimization problem.  Furthermore, the dictionary plays a  pivot role in estimation of each group sparse coefficient ${\textbf{\emph{A}}}_i$ in Eq.~\eqref{eq:2}, but the dictionary can have many different design methods. Therefore, a benchmark for sparse coding is desired to measure the sparsity of a signal/image.

Bearing the above concerns in mind, this paper provides a benchmark to sparse coding from the perspective of rank minimization. To the best of our knowledge, this is the first work to propose a benchmark to sparse coding. The contributions of our paper are as follows.
\begin{itemize}
	\item [1)] An adaptive dictionary for each patch group is devised to bridge the gap between the GSC and rank minimization models.
	\item [2)] Based on this dictionary learning scheme, we prove the equivalence of the GSC and rank minimization models, and thus the sparse coefficients of each patch group can be measured by computing the singular values of each patch group. Therefore, under this scheme, a benchmark is earned to measure the sparsity of each patch group since the singular values of the original image patch groups can be easily obtained by the singular value decomposition (SVD). Then, we can apply this benchmark to evaluate the performance of any kind of norm minimization methods in sparse coding. In other words, we can measure different norm minimization methods in sparse coding through considering their corresponding rank minimization counterparts.
	\item [3)] Four well-known rank minimization methods, \ie,  standard nuclear norm minimization (NNM) \cite{24},  Schatten $p$-norm minimization (SNM) \cite{25}, weighted nuclear norm minimization (WNNM) \cite{26} and the weighted Schatten $p$-norm minimization (WSNM) \cite{27}, are used to analyze the sparsity of each patch group\footnote{Some other rank minimization methods \cite{28,29,30,31,32,73} can also be used to analyze the sparsity of each patch group.} and the solution of WSNM is found to be the nearest one to the real singular values of each patch group, please refer to Figs.~\ref{fig:1}-\ref{fig:2} for a demonstration.
	\item [4)] Inspired by the equivalence of the rank minimization and GSC, NNM, SNM, WNNM and WSNM can be equivalently translated into $\ell_1$-norm minimization, $\ell_p$-norm minimization, weighted $\ell_1$-norm minimization and the weighted $\ell_p$-norm minimization problems in GSC, respectively.  Then, based on the earned benchmark in sparse coding, the weighted $\ell_p$-norm minimization is expected to obtain better performance than the three other norm minimization methods.
	\item [5)] To verify the feasibility of the proposed benchmark, the weighted $\ell_p$-norm minimization is used to compare with the three other norm minimization methods under the GSC framework\footnote{Note that the proposed benchmark is suitable for traditional PSC model, since the GSC is the generalized form of the PSC (when similar matched patch number $m = 1$).}. To make the proposed scheme tractable and robust, the alternating direction method of multipliers (ADMM) framework is developed to solve the non-convex weighted $\ell_p$-norm minimization problem. Experimental results on image inpainting and image compressive sensing (CS) recovery applications, show that the proposed scheme is feasible and outperforms many state-of-the-art methods in both objective and  perceptual quality.
\end{itemize}
The rest of this paper is organized as follows. Section~\ref{sec:2} briefly reviews the rank minimization methods and some nuclear norms. Section~\ref{sec:3} introduces the adaptive dictionary learning approach and presents the proposed benchmark to sparse coding from the perspective of rank minimization. Section~\ref{sec:4} develops an efficient algorithm to solve the weighted $\ell_p$-norm minimization problem based on the ADMM framework. Section~\ref{sec:5} presents experimental results and Section~\ref{sec:6} concludes the entire paper. The preliminary version of this work has appeared in~\cite{33}\footnote{Significant changes have been made compared to our previous work in \cite{33}. Specifically, we have added the analysis of the proposed benchmark using the rank minimization methods in Sec.~\ref{sec:3.3}. The gradient descent algorithm is introduced in Sec.~\ref{sec:4.1.1} to solve the CS recovery problem. Extensive experimental results have been added to verify the feasibility, robustness, and convergence of the proposed algorithm in Sec.~\ref{sec:5}.}.

\section{Background and Related Work}
\label{sec:2}
\subsection{Rank Minimization Methods}

The main goal of low-rank matrix approximation (LRMA) is to recover the underlying low-rank structure of a matrix from its degraded/corrupted observation. In general, methods of LRMA can be classified into two categories: the low-rank matrix factorization (LRMF) methods \cite{34,35,36} and the rank minimization methods \cite{24,25,26,27,28,29,30,31,32,73}. 
In this work, we focus on the {\em rank minimization} problem. To be concrete, for an input matrix $\textbf{\emph{Y}}$, the rank minimization approach aims to find a {low-rank} matrix $\textbf{\emph{X}}$, which is as close to $\textbf{\emph{Y}}$ as possible under the $F$-norm data fidelity term with a nuclear norm constraint,
\begin{equation}
\hat{\textbf{\emph{X}}}= \arg\min_{\textbf{\emph{X}}} \left(\frac{1}{2}\left\|{\textbf{\emph{Y}}}-{\textbf{\emph{X}}}\right\|_F^2+\lambda \textbf{\emph{R}}({\textbf{\emph{X}}})\right),
\label{eq:3}
\end{equation} 
where $\lambda$ is a parameter to balance the loss function and the low-rank regularization induced by the nuclear norm $\textbf{\emph{R}}(\textbf{\emph{X}})$, which will be introduced in the following subsection.

\subsection{Nuclear Norms}
Hereby, we briefly introduce four well-known nuclear norms, including standard nuclear norm \cite{24}, Schatten $p$-norm \cite{25}, weighted nuclear norm \cite{26} and the weighted Schatten $p$-norm \cite{27}.
\begin{definition}
A widely used {\bf standard nuclear norm} \cite{24} of a matrix $\textbf{\emph{X}}$ can be expressed as
\begin{equation}
\left\|\textbf{\emph{X}}\right\|_{*}=\sum\nolimits_{i=1}^{\rm min\{d, m\}}\sigma_i={\rm Tr}(({\textbf{\emph{X}}}^T{\textbf{\emph{X}}})^{\frac{1}{2}}),
\label{eq:4}
\end{equation}
where $\sigma_i$ is the $i$-th singular value of $\textbf{\emph{X}}$ and ${\rm Tr}(~)$ calculates the trace of the matrix in $(~)$.
\end{definition}

\begin{definition}
The {\bf Schatten $p$-norm} \cite{25} of a matrix  $\textbf{\emph{X}}$ can be expressed as
\begin{equation}
\left\|\textbf{\emph{X}}\right\|_{S_p}=\left(\sum\nolimits_{i=1}^{\rm  min\{d, m\}}\sigma_i^p \right)^{\frac{1}{p}}=\left({\rm Tr}(({\textbf{\emph{X}}}^T{\textbf{\emph{X}}})^{\frac{p}{2}})\right)^{\frac{1}{p}},
\label{eq:5}
\end{equation}
where $0<p\leq 1$, and $\sigma_i$ is the $i$-th singular value of $\textbf{\emph{X}}$.
\end{definition}
\begin{definition}
The {\bf weighted nuclear norm} \cite{26} of a matrix $\textbf{\emph{X}}$ can be expressed as
\begin{equation}
\left\|\textbf{\emph{X}}\right\|_{\textbf{\emph{w},*}}=\left(\sum\nolimits_{i=1}^{ \rm  min\{d, m\}}w_i\sigma_i \right)={\rm Tr}(\textbf{\emph{W}}{\boldsymbol\Sigma}),
\label{eq:6}
\end{equation}
where ${{\textbf{\emph{w}}}}=[w_1,...,w_{min\{d, m\}}]$, $\sigma_i$ is the $i$-th singular value of $\textbf{\emph{X}}$ and $w_i\geq0$ is a weight assigned to $\sigma_i$. $\textbf{\emph{W}}$ and ${\boldsymbol\Sigma}$ are diagonal matrices whose diagonal entries are composed of $w_i$ and $\sigma_i$, respectively.
\end{definition}

\begin{definition}
The {\bf weighted Schatten $p$-norm} \cite{27} of a matrix $\textbf{\emph{X}}$ is
\begin{equation}
\left\|\textbf{\emph{X}}\right\|_{\textbf{\emph{w}},S_p}=\left(\sum\nolimits_{i=1}^{ \rm min\{d, m\}}w_i\sigma_i^p\right)^{\frac{1}{p}},
\label{eq:7}
\end{equation}
where $0<p\leq 1$, and $\sigma_i$ is the $i$-th singular value of $\textbf{\emph{X}}$. $\textbf{\emph{w}}=[w_1,...,w_{min\{d, m\}}]$ and $w_i\geq0$ is a weight assigned to $\sigma_i$.

The weighted Schatten $p$-norm of $\textbf{\emph{X}}$ with {\bf power $p$} is
\begin{equation}
\left\|\textbf{\emph{X}}\right\|_{\textbf{\emph{w}},S_p}^p= \sum\nolimits_{i=1}^{\rm  min\{d, m\}}w_i\sigma_i^p ={\rm Tr}(\textbf{\emph{W}}{\boldsymbol\Sigma}^p),
\label{eq:8}
\end{equation}
where $\textbf{\emph{W}}$ and ${\boldsymbol\Sigma}$ are diagonal matrices whose diagonal entries are composed of $w_i$ and $\sigma_i$, respectively.
\end{definition}

\section{A Benchmark for Sparse Coding based on Rank Minimization Methods}
\label{sec:3}

We hereby present a benchmark to sparse coding from the perspective of rank minimization. Towards this end, an adaptive dictionary for each patch group is designed to bridge the gap between the GSC and rank minimization models. Stemmed from this dictionary learning scheme, we prove that GSC is equivalent to the rank minimization problem, and then the sparse coefficients in each patch group can be measured by calculating the singular values of each patch group. Thereby, we have a benchmark to measure the sparsity of each patch group via the rank minimization since the singular values of the original image patch groups can be easily obtained by the SVD operator\footnote{Note that the singular values have clear physical meanings in various practical problems \cite{26,59,60}. The SVD operator of an object is more able to reflect its essential property than some estimation algorithms, such as Hard-Thresholding \cite{57} and OMP \cite{58}.}. This benchmark can be used to compare the performance of any kind of norm minimization methods in sparse coding.  We can thus  measure different norm minimization methods in sparse coding through analyzing their corresponding rank minimization counterparts. In this way, we have achieved a clear visual comparison to analyze the sparsity of each patch group, please refer to  Figs.~\ref{fig:1}-\ref{fig:2} for a demonstration.

\subsection{Adaptive Dictionary Learning}
An adaptive dictionary learning approach is now designed, that is, for each patch group ${\textbf{\emph{X}}}_i$, its adaptive dictionary can be learned from its observation ${\textbf{\emph{Y}}}_i\in\mathbb{R}^{d \times m}$.
Specifically, we apply SVD to ${\textbf{\emph{Y}}}_i$,
\begin{equation}
{\textbf{\emph{Y}}}_i= {\textbf{\emph{U}}}_i{\boldsymbol\Delta}_i{\textbf{\emph{V}}}_i^T=\sum\nolimits_{j=1}^{n_1} \delta_{i,j}{\textbf{\emph{u}}}_{i,j}{\textbf{\emph{v}}}_{i,j}^T,
\label{eq:9}
\end{equation}
where ${\boldsymbol\Delta}_i={\rm diag}(\delta_{i,1}, \delta_{i,2},\dots, \delta_{i,{n_1}})$ is a diagonal matrix, ${n_1}\stackrel{\rm def}{=}{\rm min}(d,m)$,  and  ${\textbf{\emph{u}}}_{i,j}, {\textbf{\emph{v}}}_{i,j}$ are the columns of ${\textbf{\emph{U}}}_i$ and ${\textbf{\emph{V}}}_i$, respectively.

Following this, we define each dictionary atom $\textbf{\emph{d}}_{i,j}$ of the adaptive dictionary $\textbf{\emph{D}}_i$ for every patch group $\textbf{\emph{Y}}_i$, \ie,
\begin{equation}
\textbf{\emph{d}}_{i,j}={\textbf{\emph{u}}}_{i,j}{\textbf{\emph{v}}}_{i,j}^T, \ \ \ j=1,2,\dots,{n_1}.
\label{eq:10}
\end{equation}
Till now, we have learned an adaptive dictionary, \ie,
\begin{equation}
\textbf{\emph{D}}_i=[\textbf{\emph{d}}_{i,1},\textbf{\emph{d}}_{i,2},\dots,\textbf{\emph{d}}_{i,{n_1}}].
\label{eq:11}
\end{equation}
It can be seen that the designed  dictionary learning approach only requires one SVD operation per patch group.

\subsection{Prove the equivalence of Group-based Sparse Coding and Rank Minimization Problem}

In order to present the proposed benchmark to sparse coding, we first prove that GSC is equivalent to the rank minimization models based on the designed dictionary, and the following conclusions are required.
\begin{lemma}
\label{lemma:1}
The minimization problem
\begin{equation}
\hat{\textbf{{x}}}=\arg\min_{\textbf{{x}}} \left(\frac{1}{2}\left\|{\textbf{{x}}}-{\textbf{{a}}}\right\|_2^2+\tau\left\|{\textbf{{x}}}\right\|_1\right),
\label{eq:12}
\end{equation} 
has a closed-form solution
\begin{equation}
\hat{\textbf{{x}}}={\rm soft}({\textbf{{a}}},\tau)= {\rm sgn}({\textbf{{a}}})\odot{\rm max}({\rm abs}({\textbf{{a}}})-\tau,0),
\label{eq:13}
\end{equation} 
where $\odot$ denotes the element-wise (Hadamard) product.
\end{lemma}
\begin{proof}
See \cite{37}.
\end{proof}

Consider the SVD of a matrix $\textbf{\emph{Y}}\in\mathbb{R}^{d\times m}$ with rank $r$,
\begin{equation}
\textbf{\emph{Y}}= \textbf{\emph{U}}\boldsymbol\Delta\textbf{\emph{V}}^T, \quad\boldsymbol\Delta ={\rm diag}(\{\delta_i\}_{1\leq i\leq r}),
\label{eq:14}
\end{equation} 
where $\textbf{\emph{U}}\in\mathbb{R}^{d \times r}$ and $\textbf{\emph{V}}\in\mathbb{R}^{m \times r}$ are orthogonal matrices; $\delta_i$ is the $i$-th singular value of $\textbf{\emph{Y}}$. For any $\lambda\geq0$, the soft-thresholding operator $\mathcal{D}_\lambda$ is defined as
\begin{equation}
\mathcal{D}_\lambda(\textbf{\emph{Y}})= \textbf{\emph{U}} \mathcal{D}_\lambda(\boldsymbol\Delta)\textbf{\emph{V}}^T,\quad \mathcal{D}_\lambda(\boldsymbol\Delta)= {\rm soft}(\delta_i,\lambda).
\label{eq:15}
\end{equation} 
Then, we have the following Theorem.
\begin{theorem}
\label{lemma:2}
For any $\lambda\geq0$ and $\textbf{{Y}}\in\mathbb{R}^{d\times m}$, the singular value shrinkage operator in Eq.~\eqref{eq:15} satisfies
\begin{equation}
\mathcal{D}_\lambda(\textbf{{Y}})= \arg\min_{\textbf{{X}}} \left(\frac{1}{2}\left\|{\textbf{{Y}}}-{\textbf{{X}}}\right\|_F^2+\lambda\left\|{\textbf{{X}}}\right\|_*\right).
\label{eq:16}
\end{equation} 
\end{theorem}
\begin{proof}
See \cite{24}.
\end{proof}

Recalling the adaptive dictionary defined in Eq.~\eqref{eq:11}, and the classical $\ell_1$-norm minimization based GSC problem  can be represented as
\begin{equation}
\hat{\textbf{\emph{A}}}_i=\arg\min_{{\textbf{\emph{A}}}_i}  \left(\frac{1}{2}\left\|{\textbf{\emph{Y}}}_i-{\textbf{\emph{D}}}_i{\textbf{\emph{A}}}_i\right\|_F^2+\lambda\left\|{\textbf{\emph{A}}}_i\right\|_1\right).
\label{eq:17}
\end{equation} 
According to the above design of the adaptive dictionary ${\textbf{\emph{D}}}_i$  in Eq.~\eqref{eq:11}, we have the following Lemma.

\begin{lemma}
\label{lemma:3}
\begin{equation}
\left\|{\textbf{{Y}}}_i-{\textbf{{X}}}_i\right\|_F^2=\left\|{\textbf{{B}}}_i-{\textbf{{A}}}_i\right\|_F^2,
\label{eq:18}
\end{equation} 
where ${\textbf{{Y}}}_i={\textbf{{D}}}_i{\textbf{{B}}}_i$ and ${\textbf{{X}}}_i={\textbf{{D}}}_i{\textbf{{A}}}_i$.
\end{lemma}
\begin{proof}
See Appendix~\ref{lemma3}.
\end{proof}

Based on Lemmas~\ref{lemma:1}-\ref{lemma:3} and Theorem~\ref{lemma:2}, we have the following theorem.

\begin{theorem}
\label{theorem:1}
The equivalence of $\ell_1$-norm minimization based GSC and the NNM problems is satisfied under the designed adaptive dictionary ${\textbf{{D}}}_i$, i.e.,
  \begin{equation}
  \begin{aligned}
\hat{\textbf{{A}}}_i&=\arg\min\limits_{{\textbf{{A}}}_i}
\left(\frac{1}{2}\left\|{\textbf{{Y}}}_i-{\textbf{{D}}}_i{\textbf{{A}}}_i\right\|_F^2+\lambda\left\|{\textbf{{A}}}_i\right\|_1\right)\\
&\ \ \ \ \ \ \ \ \ \ \ \ \ \ \ \ \ \ \  \ \ \ \ \ \ \ \ \ \  \Updownarrow\\
\hat{\textbf{{X}}}_i&= \arg\min_{\textbf{{X}}_i}   \left(\frac{1}{2}\left\|{\textbf{{Y}}}_i-{\textbf{{X}}}_i\right\|_F^2+ \lambda \left\|{\textbf{{X}}}_i\right\|_*\right).
\end{aligned}
\label{eq:19}
\end{equation}
\end{theorem}
\begin{proof}
See Appendix~\ref{theorem1}.
\end{proof}
Similar to Theorem~\ref{theorem:1}, we have the following conclusion.

\begin{corollary}
\label{coeollary:1}
Based on the designed adaptive dictionary ${\textbf{{D}}}_i$, the weighted $\ell_1$-norm minimization based GSC, $\ell_p$-norm minimization based GSC and the weighted $\ell_p$-norm minimization based GSC problems are equivalent to WNNM, SNM and the WSNM problems, respectively.
\end{corollary}
\begin{proof}
See Appendix~\ref{coeolllar}.
\end{proof}

Note that the dictionary can be learned in various manners and the designed adaptive dictionary learning approach is just one of them. The above conclusions cannot hold for some general dictionaries \cite{3,4,38,39}. Therefore, to further demonstrate the universality and advantage of the proposed scheme, another two generalized dictionary learning methods are also exploited to verify the feasibility of the proposed benchmark, \ie, graph-based dictionary learning method \cite{38} and  PCA dictionary learning method \cite{39} (See subsection~\ref{5.2} for more details).

\begin{figure}[!t]
		\centering
\begin{minipage}[b]{1\linewidth}
{\includegraphics[width= 1\textwidth]{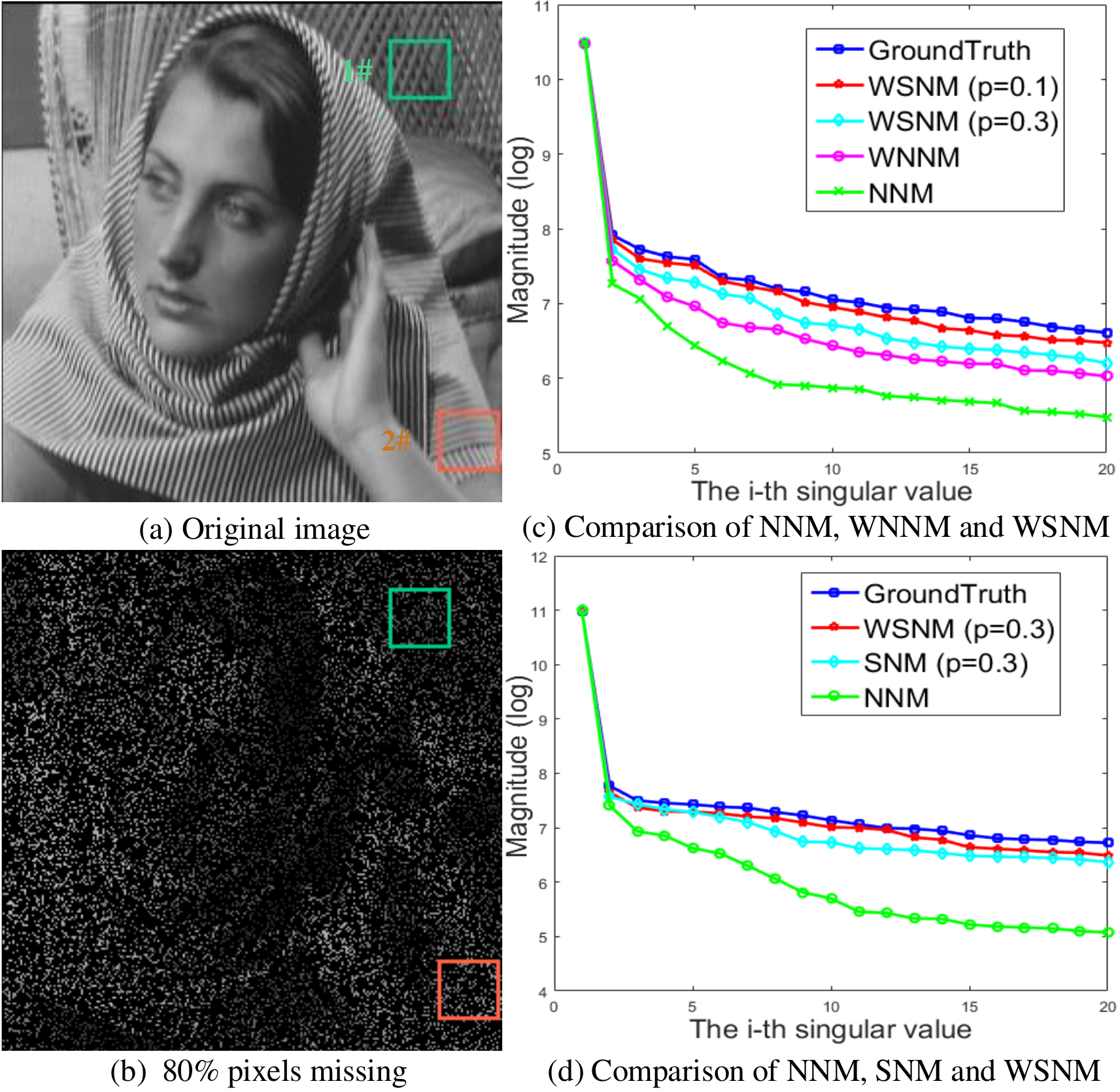}}
\end{minipage}
\vspace{-4mm}
	\caption{Analyzing the sparsity of each patch group based on the  rank  minimization scheme in terms of image inpainting. (a) Original {\em Barbara} image. (b) 80\% pixels are missing. (c-d) The curved lines of the singular values using different rank minimization methods of the patch group with reference in the cyan (1\#) and orange (2\#) boxes, respectively.}
	\vspace{-2mm}
	\label{fig:1}
\end{figure}

\subsection{A Benchmark for Sparse Coding based on the Rank Minimization Methods\label{sec:3.3}}
Based on  Theorem~\ref{theorem:1} and Corollary~\ref{coeollary:1}, the GSC problem can be transformed into the rank minimization problem.
%
We therefore possess a benchmark to measure the sparsity of each patch group via the rank minimization scheme because the singular values of the original image patch groups can be easily obtained by the SVD operator. 
Then, we can utilize the proposed benchmark to measure different norm minimization methods in sparse coding through analyzing their corresponding rank minimization methods.

\begin{figure}[!t]
		\centering
\begin{minipage}[b]{1\linewidth}
{\includegraphics[width= 1\textwidth]{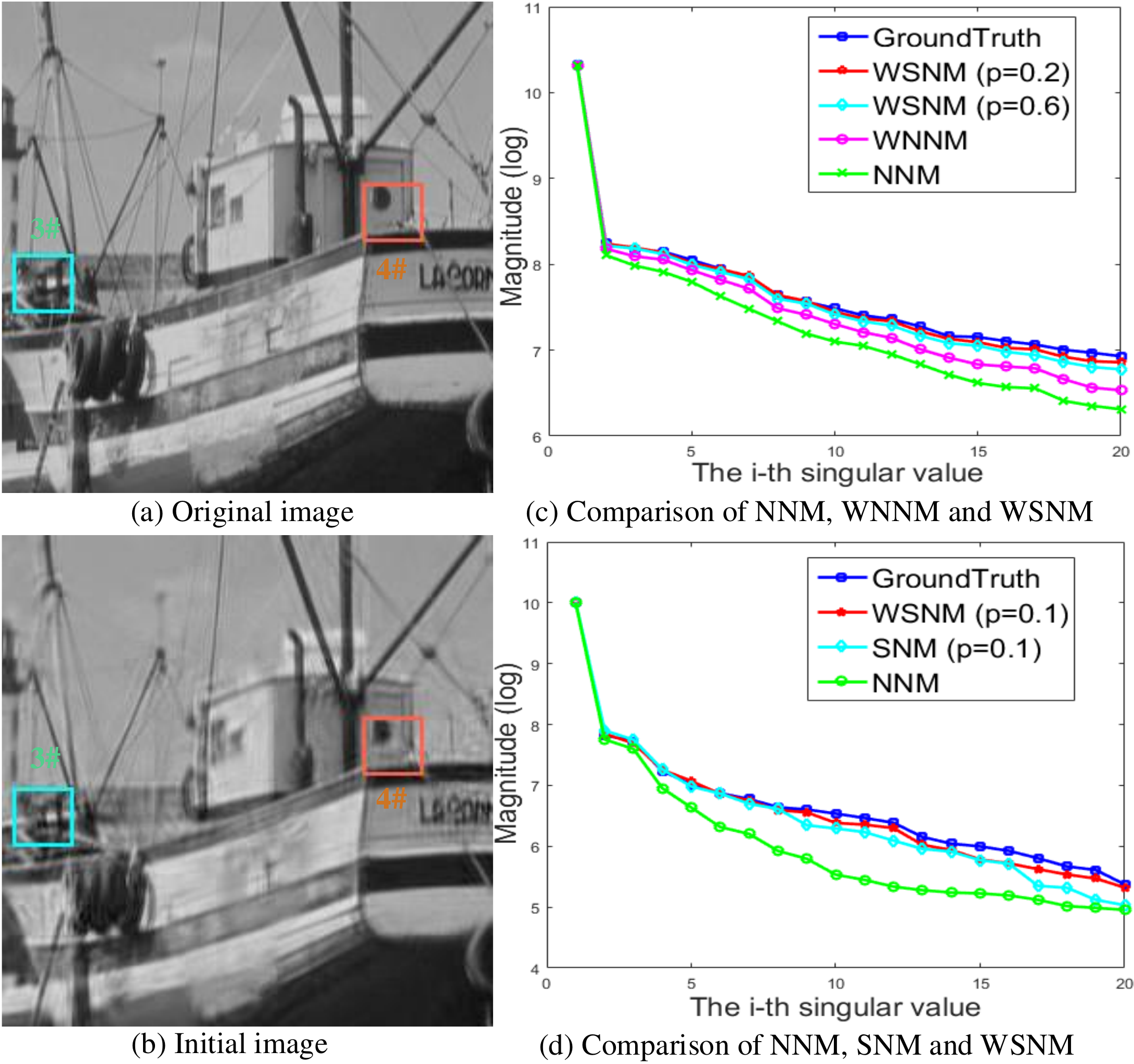}}
\end{minipage}
\vspace{-4mm}
	\caption{Analyzing the sparsity of each patch group based on the rank minimization scheme in terms of image CS recovery. The image \emph{boats} in (a) is compressively sampled by a random Gaussian matrix with 0.2$N$ measurements, and an initial image in (b) is estimated by using the BCS based CS image recovery method \cite{40}. (c-d) The curved lines of the singular values using different rank minimization methods of the patch group with reference in the cyan (3\#) and orange (4\#) boxes, respectively.}
	\label{fig:2}
	\vspace{-2mm}
\end{figure}

Specifically, four well-known rank minimization methods are used to constrain Eq.~\eqref{eq:3} to study the sparsity of each patch group, \ie, NNM, SNM, WNNM and WSNM. In these experiments, two widely used images, namely, \emph{Barbara} and \emph{boats}, are used as examples in the context of image inpainting and image CS recovery, respectively. In image inpainting, 80\% pixels of image \emph{Barbara} are damaged in Fig.~\ref{fig:1}(b) and two patch groups based on the 1\# position and the 2\# position are generated in Fig.~\ref{fig:1}(a). In image CS recovery, image \emph{boats} is compressively sampled by a random Gaussian matrix with 0.2$N$ measurements and an initial image is estimated by using a standard CS recovery method (\emph{e.g.}, DCT/BCS \cite{40} based CS reconstruction method) shown in Fig.~\ref{fig:2}(b). We conduct two patch groups based on the 3\# position and the 4\# position in Fig.~\ref{fig:2}(a).  From Fig.~\ref{fig:1}(c-d) and Fig.~\ref{fig:2}(c-d), we can clearly observe that the singular values of  WSNM results are the best approximation to the ground-truth in comparison with the three other rank minimization methods, \ie, NNM, SNM and WNNM.

Based on Theorem~\ref{theorem:1} and Corollary~\ref{coeollary:1}, NNM, SNM, WNNM and WSNM can be equivalently transformed to $\ell_1$-norm minimization, $\ell_p$-norm minimization, weighted $\ell_1$-norm minimization and the weighted $\ell_p$-norm minimization problems in GSC, respectively. Meanwhile, based on the earned benchmark, \ie, according to the above analysis of Fig.~\ref{fig:1} and Fig.~\ref{fig:2}, we also wish that the weighted $\ell_p$-norm minimization can achieve better performance than the three other norm minimization methods in sparse coding, including $\ell_1$-norm minimization, $\ell_p$-norm minimization and the weighted $\ell_1$-norm minimization. To verify the feasibility of the proposed benchmark, we compare the weighted $\ell_p$-norm minimization with the three other norm minimization methods in sparse coding. Specifically, based on the GSC framework, we exploit these  methods to solve image restoration tasks including image inpainting and image CS recovery.

\section{Validation of the Benchmark for Sparse Coding Via the Weighted $\ell_p$-Norm Minimization}
\label{sec:4}

According to the above analysis and based on our proposed benchmark in sparse coding, the weighted $\ell_p$-norm minimization is expected to achieve better results than the three other norm minimization methods in sparse coding.

In this section, we apply two image restoration tasks including image inpainting and image CS recovery to verify the feasibility of the proposed benchmark. Specifically, the goal of image restoration \cite{74,76,41,42,43,44,45,46,47,48,49,50,77} is to reconstruct a high quality image $\textbf{\emph{x}}$ from its degraded observation $\textbf{\emph{y}}$,
\begin{equation}
\textbf{\emph{y}}=\textbf{\emph{H}}\textbf{\emph{x}} +  \textbf{\emph{n}},
\label{eq:20}
\end{equation} 
where $\textbf{\emph{H}}$ is a non-invertible linear degradation operator and $\textbf{\emph{n}}$ is usually assumed to be a zero-mean white Gaussian noise. With different setting of matrix $\textbf{\emph{H}}$, Eq.~\eqref{eq:20} can represent different image restoration tasks. For instance, when  $\textbf{\emph{H}}$ is a diagonal matrix whose diagonal entries are either 1 or 0, keeping or killing corresponding pixels, Eq.~\eqref{eq:20} becomes image inpainting \cite{41,42,43,44,45}; when $\textbf{\emph{H}}$ is a random projection matrix with more columns than rows, Eq.~\eqref{eq:20} becomes image CS recovery \cite{46,47,48,49,50}. 

Then, we exploit the case of the weighted $\ell_p$-norm minimization as an example. Given the degraded image $\textbf{\emph{y}}$ in Eq.~\eqref{eq:20}, we aim to recover the original image $\textbf{\emph{x}}$ by solving the following non-convex weighted $\ell_p$-norm minimization based sparse coding problem,
  \begin{equation}
\hat {\boldsymbol\alpha}=\arg\min\limits_{{\boldsymbol\alpha}}
\frac{1}{2}\left\|{\textbf{\emph{y}}}-\textbf{\emph{H}}{\textbf{\emph{D}}}{\boldsymbol\alpha}\right\|_2^2
+\lambda\left\|{\textbf{\emph{w}}}{\boldsymbol\alpha}\right\|_p,
\label{eq:21}
\end{equation} 
where $\textbf{\emph{D}}$ represents the dictionary and $\boldsymbol \alpha$ is the sparse coefficient; ${\textbf{\emph{w}}}$ is a weight in the weighted $\ell_p$-norm and $\lambda$ is a regularization parameter.

\subsection{ADMM based Algorithm for the Weighted $\ell_p$-norm Minimization}

Solving the objective function of Eq.~\eqref{eq:21} is difficult, since it is a large scale non-convex optimization problem. In order to make the optimization problem tractable, we employ the alternating direction method of multipliers (ADMM) \cite{51,52} to solve Eq.~\eqref{eq:21}. Numerical simulations have shown that ADMM can converge by only using a small memory footprint, which makes it attractive for various large-scale optimization problems \cite{53,54,55}.

Now, let us come back to Eq.~\eqref{eq:21} and use ADMM to solve it. We first turn Eq.~\eqref{eq:21} into another equivalent constrained form by introducing an auxiliary variable $\textbf{\emph{z}}$,
\begin{equation}
\hat {\boldsymbol\alpha}=\arg\min\limits_{\textbf{\emph{z}}, {\boldsymbol\alpha}}
\frac{1}{2}\left\|{\textbf{\emph{y}}}-\textbf{\emph{H}}{\textbf{\emph{z}}}\right\|_2^2+\lambda \left\|{\textbf{\emph{w}}}\boldsymbol\alpha\right\|_p,\ {\rm s.t.}\ \  \textbf{\emph{z}}={\textbf{\emph{D}}}{\boldsymbol\alpha}.
\label{eq:23}
\end{equation} 
Following this,  Eq.~\eqref{eq:23} can be transformed into three iterative steps:
\begin{align}
{\textbf{\emph{z}}}^{t+1}&=\arg\min\limits_{\textbf{\emph{z}}}\frac{1}{2}\left\|\textbf{\emph{y}}-\textbf{\emph{H}}{\textbf{\emph{z}}}\right\|_2^2
+\frac{\rho}{2}\left\|\textbf{\emph{z}}-{{\textbf{\emph{D}}}{\boldsymbol\alpha}}^{t}-\textbf{\emph{b}}^{t}\right\|_2^2
\label{eq:24},\\
{\boldsymbol\alpha}^{t+1}&=\arg\min\limits_{\boldsymbol\alpha}\lambda\left\|{\textbf{\emph{w}}}{\boldsymbol\alpha}\right\|_p
+\frac{\rho}{2}\left\|\textbf{\emph{z}}^{t+1}-{{\textbf{\emph{D}}}{\boldsymbol\alpha}}-\textbf{\emph{b}}^{t}\right\|_2^2
\label{eq:25},\\
{\textbf{\emph{b}}}^{t+1}&={\textbf{\emph{b}}}^{t}-({\textbf{\emph{z}}}^{t+1}-{{\textbf{\emph{D}}}{\boldsymbol\alpha}}^{t+1})
\label{eq:26}.
\end{align}
One can observe that the minimization of Eq.~\eqref{eq:23} is separated into two minimization sub-problems, \ie,~$\textbf{\emph{z}}$ and $\boldsymbol\alpha$ sub-problem. Fortunately, there is an efficient solution to each sub-problem, which will be discussed in the following subsections.
The superscript $t$ is omitted for conciseness in the derivation below.

\subsubsection{${\textbf{\emph{z}}}$ Sub-problem \label{sec:4.1.1}}
Given $\boldsymbol\alpha$,  $\textbf{\emph{z}}$ sub-problem in Eq.~\eqref{eq:24} becomes
\begin{equation}
\min_{\textbf{\emph{z}}}{\textbf{\emph{Q}}}_1({{\textbf{\emph{z}}}})=\min\limits_{\textbf{\emph{z}}}\frac{1}{2}\left\|\textbf{\emph{y}}-\textbf{\emph{H}}{\textbf{\emph{z}}}\right\|_2^2
+\frac{\rho}{2}\left\|\textbf{\emph{z}}-{{\textbf{\emph{D}}}{\boldsymbol\alpha}}-\textbf{\emph{b}}\right\|_2^2.
\label{eq:27}
\end{equation}
This is a quadratic form and  has a closed-form solution,
\begin{equation}
\hat{\textbf{\emph{z}}}=\left({\textbf{\emph{H}}}^{T}{\textbf{\emph{H}}}+\rho{\textbf{\emph{I}}}\right)^{-1}\left({\textbf{\emph{H}}}^{T}{\textbf{\emph{y}}}
+\rho({{\textbf{\emph{D}}}{\boldsymbol\alpha}}+\textbf{\emph{b}})\right),
\label{eq:28}
\end{equation} 
where ${\textbf{\emph{I}}}$ is an identity matrix with the desired dimension.

Owing to the specific structure of ${\textbf{\emph{H}}}$ in image inpainting, Eq.~\eqref{eq:27} can be efficiently computed without matrix inversion. However, in image CS recovery, as ${\textbf{\emph{H}}}$ is a random projection matrix, it is expensive to solve Eq.~\eqref{eq:28} directly. In this work, we employ the steepest descent method \cite{56} to solve Eq.~\eqref{eq:27},
\begin{equation}
\hat{\textbf{\emph{z}}}={\textbf{\emph{z}}}-\eta{\textbf{\emph{q}}},
\label{eq:29}
\end{equation} 
where ${\textbf{\emph{q}}}$ is the gradient of the objective function ${\textbf{\emph{Q}}}_1({{\textbf{\emph{z}}}})$, and {$\eta$ represents the  step size.}

Thereby, in image CS recovery, we only need an iterative calculation to solve  ${\textbf{\emph{z}}}$ sub-problem, \ie,
\begin{equation}
\hat{\textbf{\emph{z}}}={\textbf{\emph{z}}}-\eta\left({{\textbf{\emph{H}}}}^{T}{{\textbf{\emph{H}}}}{\textbf{\emph{z}}}
-{{\textbf{\emph{H}}}}^{T}{\textbf{\emph{y}}}+\rho(\textbf{\emph{z}}-{{\textbf{\emph{D}}}{\boldsymbol\alpha}}-\textbf{\emph{b}})\right),
\label{eq:30}
\end{equation}
where ${{\textbf{\emph{H}}}}^{T}{{\textbf{\emph{H}}}}$ and ${{\textbf{\emph{H}}}}^{T}{\textbf{\emph{y}}}$ can be pre-calculated.

\subsubsection{$\boldsymbol\alpha$ Sub-problem}

Given ${\textbf{\emph{z}}}$,  $\boldsymbol\alpha$ sub-problem in Eq.~\eqref{eq:25} can be rewritten as
\begin{equation}
\min_{\boldsymbol\alpha}{\textbf{\emph{Q}}}_2({\boldsymbol\alpha})=
\min\limits_{\boldsymbol\alpha}
\frac{1}{2}\left\|{{\textbf{\emph{D}}}\boldsymbol\alpha}-\textbf{\emph{l}}\right\|_2^2\\
+\frac{\lambda}{\rho}\left\|{\textbf{\emph{w}}}\boldsymbol\alpha\right\|_p,
\label{eq:31}
\end{equation}
where  $\textbf{\emph{l}}=\textbf{\emph{z}}-\textbf{\emph{b}}$.

However, due to the complicated structure of $\|{\textbf{\emph{w}}}{\boldsymbol\alpha}\|_p$, it is challenging to solve Eq.~\eqref{eq:31} directly. Let $\textbf{\emph{x}}={{\textbf{\emph{D}}}{\boldsymbol\alpha}}$, Eq.~\eqref{eq:31} can be rewritten as
\begin{equation}
\min_{\boldsymbol\alpha}{\textbf{\emph{Q}}}_2({\boldsymbol\alpha})=
\min\limits_{\boldsymbol\alpha}
\frac{1}{2}\left\|{{\textbf{\emph{x}}}}-\textbf{\emph{l}}\right\|_2^2\\
+\frac{\lambda}{\rho}\left\|{\textbf{\emph{w}}}{\boldsymbol\alpha}\right\|_p.
\label{eq:32}
\end{equation}
Since the basic unit of the GSC model is {patch group} and in order to achieve a tractable solution to Eq.~\eqref{eq:32}, a general assumption is made, and with this even a closed-form solution can be achieved. Specifically, $\textbf{\emph{l}}$ can be regarded as a noisy observation of $\textbf{\emph{x}}$, and then the assumption is made that each element of
$\textbf{\emph{e}}=\textbf{\emph{x}}-\textbf{\emph{l}}$ follows an independent zero-mean Gaussian distribution with variance
${\sigma}_n^{2}$. Provided this assumption, we have the following theorem.

\begin{theorem}
\label{theorem:2}
Define $\textbf{{x}}, \textbf{{l}}\in\mathbb{R}^{N}$, ${\textbf{{X}}}_i$, ${\textbf{{L}}}_i \in\mathbb{R}^{d\times m}$, and $e_j$  denotes the $j$-th element of the error vector ${\textbf{{e}}}\in\mathbb{R}^{N}$, where $\textbf{{e}}=\textbf{{x}}-\textbf{{l}}$. Assume that $e_j$ follows an independent zero mean Gaussian distribution with variance ${\sigma}_n^{2}$, and thus for any $\xi>0$, we can represent the relationship between $\frac{1}{N}\left\|\textbf{{x}}-\textbf{{l}}\right\|_2^2$ and ${\frac{1}{S}}\sum_{i=1}^n\left\|{\textbf{{X}}}_i-{\textbf{{L}}}_i\right\|_F^2$  by the following property,
\begin{equation}	\lim_{{N\rightarrow\infty}\atop{S\rightarrow\infty}}{\textbf{{P}}}{\left(\left|\frac{1}{N}\left\|\textbf{{x}}-\textbf{{l}}\right\|_2^2
-{\frac{1}{S}}\sum\nolimits_{i=1}^n\left\|{\textbf{{X}}}_i-{\textbf{{L}}}_i\right\|_F^2\right|<\xi\right)}=1,
\label{eq:33}
\end{equation}
where ${\textbf{{P}}}(~)$  represents the probability and $S=d\times m \times n$.
\end{theorem}
\begin{proof}
See \cite{6}.
\end{proof}	

Then, based on Theorem \ref{theorem:2}, we have the following equation with a very large probability (limited to 1) at each iteration,
\begin{equation}
\frac{1}{N}\left\|\textbf{\emph{x}}-\textbf{\emph{l}}\right\|_2^2
={\frac{1}{S}}\sum\nolimits_{i=1}^n\left\|{\textbf{\emph{X}}}_i-{\textbf{\emph{L}}}_i\right\|_F^2.
\label{eq:34}
\end{equation}

	\begin{algorithm}[!t]
		\caption{ADMM for weighted $\ell_p$-norm minimization.}
		\begin{algorithmic}[1]
			\REQUIRE The observed image $\textbf{\emph{y}}$ and the measurement matrix $\textbf{\emph{H}}$.
			\STATE  Set parameters $t$, ${{\textbf{\emph{b}}}}$,
			${{\textbf{\emph{z}}}}$, ${{\boldsymbol\alpha}}$,
			$\emph{C}$,  $\emph{d}$, $\emph{m}$, $\rho$, $p$, $\sigma_n$, $\epsilon$ and $\varepsilon$.
			\FOR{$t=0$ \TO Max-Iter }
			\IF {$\textbf{\emph{H}}$ is mask operator}
			\STATE Update ${{\textbf{\emph{z}}}}^{t+1}$ by Eq.~\eqref{eq:28};
			\ELSIF{$\textbf{\emph{H}}$ is random projection operator}
			\STATE Update ${{\textbf{\emph{z}}}}^{t+1}$ by Eq.~\eqref{eq:30};
			\ENDIF
			\STATE ${\textbf{\emph{l}}}^{t+1} = {\textbf{\emph{z}}}^{t+1}- {\textbf{\emph{b}}}^{t}$;
			\FOR{Each group ${\textbf{\emph{L}}}_{i}$}	
			\STATE Construct dictionary ${\textbf{\emph{D}}}_i$ by computing Eq.~\eqref{eq:11};
			\STATE Update ${\lambda}^{t+1}$ by computing  Eq.~\eqref{eq:41};
			\STATE  Update $\tau^{t+1}$ by computing ${{\tau}} =\frac{{{\lambda}}{\emph{S}}}{\rho{\emph{N}}}$;
			\STATE Update ${\textbf{\emph{w}}}_i^{t+1}$  by computing  Eq.~\eqref{eq:40};
			\STATE Update ${{\boldsymbol\alpha}_i}^{t+1}$ by computing Eq.~\eqref{eq:39};
			\ENDFOR
			\STATE Update ${\textbf{\emph{D}}}^{t+1}$ by concatenating all ${\textbf{\emph{D}}}_i$.
			\STATE  Update ${\boldsymbol\alpha}^{t+1}$ by concatenating all ${\boldsymbol\alpha}_i$.
			\STATE Update ${{\textbf{\emph{b}}}^{t+1}}$ by computing Eq.~\eqref{eq:26}.
			\ENDFOR
			\STATE $\textbf{Output:}$ The final restored image $\hat{\textbf{\emph{x}}}={\textbf{\emph{D}}}{\boldsymbol\alpha}.$
		\end{algorithmic}
		\label{algo:3}
	\end{algorithm}

Therefore, Based on Eqs.~\eqref{eq:32} and ~\eqref{eq:34},  we have
\begin{equation}
\begin{aligned}
&\min\limits_{\boldsymbol\alpha}\frac{1}{2}{\left\|\textbf{\emph{x}}-\textbf{\emph{l}}\right\|_2^2}
 +\frac{\lambda}{\rho}\left\|{\textbf{\emph{w}}}{\boldsymbol\alpha}\right\|_p\\
&=\min\limits_{\{{{\textbf{\emph{A}}}}_i\}_{{i=1}}^n} \sum\nolimits_{i=1}^n\left(\frac{1}{2}\left\|{\textbf{\emph{X}}}_i-{\textbf{\emph{L}}}_i\right\|_F^2
 +{{\tau}}\left\|{\textbf{\emph{W}}}_i\circ{{\textbf{\emph{A}}}}_i\right\|_p\right) \\
 &=\min\limits_{\{{{\textbf{\emph{A}}}}_i\}_{{i=1}}^n} \sum\nolimits_{i=1}^n\left(\frac{1}{2}\left\|{\textbf{\emph{L}}}_i-{{\textbf{\emph{D}}}_i{{\textbf{\emph{A}}}}_i}\right\|_F^2
 +{{\tau}}\left\|{\textbf{\emph{W}}}_i\circ{{\textbf{\emph{A}}}}_i\right\|_p\right),
\end{aligned}
\label{eq:35}
\end{equation}
where $\circ$ represents the element-wise product of two matrices (Hadamard product); ${\textbf{\emph{X}}}_i={{\textbf{\emph{D}}}_i{\textbf{\emph{A}}}_i}$ and ${{\tau}}=\frac{{{\lambda}}{\emph{S}}}{\rho{\emph{N}}}$. Clearly, Eq.~\eqref{eq:35} can be viewed as the GSC problem by solving $n$ sub-problems for each patch group ${\textbf{\emph{X}}}_i$ \cite{5,6,7,8}.  ${{\textbf{\emph{A}}}}_i$ represents the group sparse coefficient of each patch group ${\textbf{\emph{X}}}_i$, and ${{\textbf{\emph{W}}}_i}$ is a weight assigned to each patch group ${\textbf{\emph{X}}}_i$. Each weight matrix ${{\textbf{\emph{W}}}_i}$ will enhance the representation capability of each group sparse coefficient ${{\textbf{\emph{A}}}}_i$.  
Note that here we abuse the $p$-norm for matrix and the $\ell_p$-norm is imposed on each column of group sparse coefficient ${\textbf{\emph{A}}}_i$. Based on Lemma~\ref{lemma:3}, Eq.~\eqref{eq:35} can be rewritten as
\begin{equation}
\begin{aligned}
{\{\hat{\textbf{\emph{A}}}}_i\}_{i=1}^n&=\min\limits_{\{{{\textbf{\emph{A}}}}_i\}_{i=1}^n}\sum\nolimits_{i=1}^n\left(\frac{1}{2}\left\|{\textbf{\emph{R}}}_i-{\textbf{\emph{A}}}_i\right\|_F^2
 +{{\tau}}\left\|{\textbf{\emph{W}}}_i\circ{{\textbf{\emph{A}}}}_i\right\|_p\right)\\
 &=\min\limits_{\{{\boldsymbol\alpha}_i\}_{i=1}^n}\sum\nolimits_{i=1}^n\left(\frac{1}{2}\left\|{{\boldsymbol\gamma}_i}-{{\boldsymbol\alpha}_i}\right\|_2^2
 +{{\tau}}\left\|{\textbf{\emph{w}}}_i{{\boldsymbol\alpha}_i}\right\|_p\right),
\end{aligned}
\label{eq:36}
\end{equation}
where ${\textbf{\emph{L}}}_i={{\textbf{\emph{D}}}_i{\textbf{\emph{R}}}_i}$; ${{{{\boldsymbol\alpha}}}_i}$, ${{{{\boldsymbol\gamma}}}_i}$ and ${{\textbf{\emph{w}}}}_i$ denote the vectorization of the matrix ${{{\textbf{\emph{A}}}}_i}$, ${{{\textbf{\emph{R}}}}_i}$ and ${\textbf{\emph{W}}}_i$, respectively.

Therefore, the minimization problem of Eq.~\eqref{eq:31} can be simplified to solve the minimization problem in Eq.~\eqref{eq:36}.  To obtain the solution of Eq.~\eqref{eq:36} effectively, we develop a generalized soft-thresholding (GST) algorithm \cite{61} to solve it. Specifically, a closed-form solution of Eq.~\eqref{eq:36} can be computed by
 \begin{equation}
{{\boldsymbol\alpha}_{i}}={{\emph{GST}}}(\boldsymbol\gamma_{i}, \tau {\textbf{\emph{w}}}_i, p).
\label{eq:39}
\end{equation}
Please refer to \cite{61} for more details about the GST algorithm.


\vspace{-3mm}
\subsection{Setting the Weight and Regularization Parameter}
As large values in coefficient ${{{\boldsymbol\alpha}}}_i$ usually depict major edge and texture information \cite{17}, in order to reconstruct ${\textbf{\emph{X}}}_i$ from its degraded one, we usually shrink the larger values less, while shrinking smaller ones more in each iteration \cite{26}. Therefore, we let
\begin{equation}
{\textbf{{\emph{w}}}}_i= \frac{1}{|{{{\boldsymbol\gamma}}}_{i}|+\epsilon},
\label{eq:40}
\end{equation}
with $\epsilon$ as a small positive constant.

The regularization parameter $\lambda$ that balances the fidelity term and the regularization term can be adaptively determined for better reconstruction performance. Inspired by \cite{62}, $\lambda$ of each patch group ${\textbf{\emph{L}}}_i$  in each iteration is set to:
\begin{equation}
\lambda=\frac{2\sqrt{2}\sigma_n^2}{\varphi_i+\varepsilon}.
\label{eq:41}
\end{equation}
where $\varphi_i$ denotes the estimated variance of ${{{\boldsymbol\gamma}}}_i$, and $\varepsilon$ is a small positive constant.

After solving the two sub-problems, we summarize the entire algorithm for Eq.~\eqref{eq:21} in Algorithm~\ref{algo:3}.

\begin{figure}[!t]
	\centering
\begin{minipage}[b]{1\linewidth}
{\includegraphics[width= 1\textwidth]{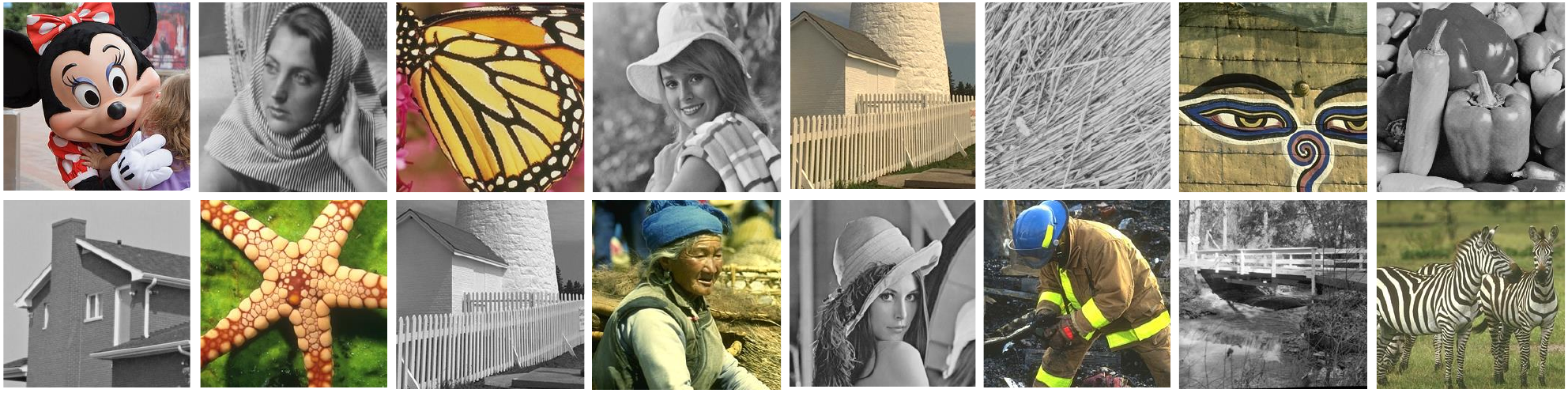}}
\end{minipage}
	\vspace{-4mm}
	\caption{Test images. Top row, from left to right: Mickey, Barbara, Butterfly, Elaine, Fence, Straw, Mural, Peppers. Bottom row, from left to right: House, Starfish, Fence, Nanna, Lena, Fireman, Bridge, Zebra.}
	\label{fig:3}
	\vspace{-2mm}
\end{figure}

\section{Experimental Results}
\label{sec:5}
In this section, extensive experiments are presented to validate the feasibility of the proposed benchmark, \ie, demonstrate that the results of the weighted $\ell_p$-norm ($w\ell_p$-norm) minimization in sparse coding are consistent with the ones of our previous analysis (WSNM) in Section~\ref{sec:3.3}. We compare the $w\ell_p$-norm minimization with the three other well-known norm minimization methods based on the GSC framework, including $\ell_1$-norm minimization, weighted $\ell_1$-norm ($w\ell_1$-norm) minimization and the $\ell_p$-norm minimization. We conduct the performance evaluations on image inpainting and image CS recovery.  The peak signal-to-noise ratio (PSNR) metric is adopted to evaluate the quality of the restored images. {We have also calculated the full reference image quality assessment (FR-IQA) metrics \cite{68,69} of the restored images, \ie, FSIM \cite{68},  which shows similar results to PSNR  and thus omitted here  due to the limited space.} The experimental test images are shown in Fig.~\ref{fig:3}. Due to limited  space, please enlarge the tables and figures on the screen for better comparison.  The source code of the proposed method is available at: {\url{https://drive.google.com/open?id=1J3cgJF2zrvcyL08COy4eexwNk-wErUv1}}.

\subsection{Parameter Selection}
\label{5.1}
The parameters used in the algorithm are empirically chosen according to different applications in order to achieve relatively good results. Note that all norm minimization methods are based on the designed adaptive dictionary learning (ADL) scheme under the GSC framework.

\begin{table}[!t]
\vspace{-3mm}
	\caption{Detailed setting of $\rho, \lambda$ for image inpainting and image CS recovery with the designed adaptive dictionary learning approach.}
	\centering
	\scriptsize
	\begin{tabular}{|c||c|c||c|c||c||c|c|c|}
		\hline
		\multicolumn{7}{|c|}{Image Inpainting}\\
		\hline
		\multicolumn{1}{|c||}{ADL}&\multicolumn{2}{|c||}{$\ell_1$-norm}&\multicolumn{2}{|c||}{$\ell_p$-norm}&\multicolumn{1}{|c||}{$w\ell_1$-norm}&\multicolumn{1}{|c|}{$w\ell_p$-norm}\\
		\hline
		\multirow{1}{*}{{{Parameters}}}&\multirow{1}{*}{{{$\rho$}}}&\multirow{1}{*}{{{$\lambda$}}}&\multirow{1}{*}{{{$\rho$}}}
		&{{{$\lambda$}}}&\multirow{1}{*}{{{$\rho$}}}&\multirow{1}{*}{{{$\rho$}}}\\
		\hline
		\multirow{1}{*}{80\%}     &    7e-5    &  5e-6      & 0.006      & 0.07   & 0.1      & 0.0003    \\
		\hline
		\multirow{1}{*}{70\%}     &    1e-4    &  7e-6      & 0.008      & 0.07   & 0.1      & 0.0003    \\
		\hline
		\multirow{1}{*}{60\%}     &    1e-5    &  1e-6      & 7e-5       & 3e-6   & 0.1      & 0.03    \\
		\hline
		\multirow{1}{*}{50\%}     &    5e-5    &  1e-5      & 0.0001       & 7e-6   & 0.1      & 0.04    \\
		\hline
		\multicolumn{7}{|c|}{Image CS Recovery}\\
		\hline
		\multicolumn{1}{|c||}{ADL}&\multicolumn{2}{|c||}{$\ell_1$-norm}&\multicolumn{2}{|c||}{$\ell_p$-norm}&\multicolumn{1}{|c||}{$w\ell_1$-norm}&\multicolumn{1}{|c|}{$w\ell_p$-norm}\\
		\hline
		\multirow{1}{*}{{{Parameters}}}&\multirow{1}{*}{{{$\rho$}}}&\multirow{1}{*}{{{$\lambda$}}}&\multirow{1}{*}{{{$\rho$}}}
		&{{{$\lambda$}}}&\multirow{1}{*}{{{$\rho$}}}&\multirow{1}{*}{{{$\rho$}}}\\
		\hline
		\multirow{1}{*}{0.2$N$}     &    0.001    &  5e-5      & 0.003      & 5e-4   & 0.1     & 0.0005    \\
		\hline
		\multirow{1}{*}{0.3$N$}     &    0.003    &  7e-4      & 0.01       & 3e-4   & 0.1     & 0.05    \\
		\hline
		\multirow{1}{*}{0.4$N$}     &    0.003    &  5e-4      & 0.006      & 3e-4   & 0.1     & 0.05    \\
		\hline
		\multirow{1}{*}{0.5$N$}     &    0.003    &  5e-4      & 0.006      & 3e-4   & 0.2     & 0.05    \\
		\hline
	\end{tabular}
	\label{lab:1}
	\vspace{-3mm}
\end{table}

In image inpainting, the mask is generated randomly. The size of patch is set to $8\times 8$. The similar patch number $m$ is set to 60. The searching window $C \times C$ is set to $25\times 25$; $\sigma_n =\sqrt{2}$; $p$ is set to 0.45, 0.45, 0.95 and 0.95 when 80\%, 70\%, 60\% and 50\% pixels are missing, respectively.  $\epsilon$ and $\varepsilon$ are set as (0.35, 0.35)  and (0.1, 0.3) for $w\ell_1$-norm and $w\ell_p$-norm, respectively. The detailed settings of $\rho$ and $\lambda$ are shown on the upper part of Table~\ref{lab:1}. Due to the existence of the weight $\textbf{\emph{w}}$, $\lambda$ is computed by Eq.~\eqref{eq:41} in $w\ell_1$-norm and $w\ell_p$-norm. We will give a detailed discussion on how to choose the best power $p$ in Sec.~\ref{5.4}.

In image CS recovery, we generate the CS measurements at the block level by utilizing a Gaussian random projection matrix to test images, \ie, CS with block size $32\times 32$ \cite{40}. The patch size is set to $7\times 7$. The similar patch number $m =60$, and the search window $C \times C$  is set to $20\times 20$; $\sigma_n =\sqrt{2}$;  $p$ is set to 0.5, 0.95, 0.95 and 0.95 with 0.2$N$, 0.3$N$, 0.4$N$ and 0.5$N$ measurements, respectively. $\epsilon$ and $\varepsilon$ are set as (0.35, 0.35)  and (0.1, 0.4) for $w\ell_1$-norm and $w\ell_p$-norm, respectively. Similarly, $\lambda$ is computed by Eq.~\eqref{eq:41} in $w\ell_1$-norm and $w\ell_p$-norm. The detailed settings of $\rho$ and $\lambda$ are shown at the lower part of Table~\ref{lab:1}.

In addition, to make a fair comparison of all norm minimization methods, the iterative stopping criterion is set to: PSNR ($t+1$) - PSNR ($t$) $<$ 0, where PSNR ($t+1$) and PSNR ($t$) denote the PSNR values of the restored images at the $(t+1)$-th iteration and $t$-th iteration, respectively.

\subsection{Comparisons of $\ell_1$-norm, Weighted $\ell_1$-norm, $\ell_p$-norm and the Weighted $\ell_p$-norm}
\label{5.2}
We first compare four norm minimization methods, \ie, $\ell_1$-norm minimization, $w \ell_1$-norm minimization, $\ell_p$-norm minimization and the $w \ell_p$-norm minimization,  based on the designed ADL scheme for image inpainting and image CS recovery.

\begin{figure}[!t]
		 \centering
\begin{minipage}[b]{1\linewidth}
{\includegraphics[width= 1\textwidth]{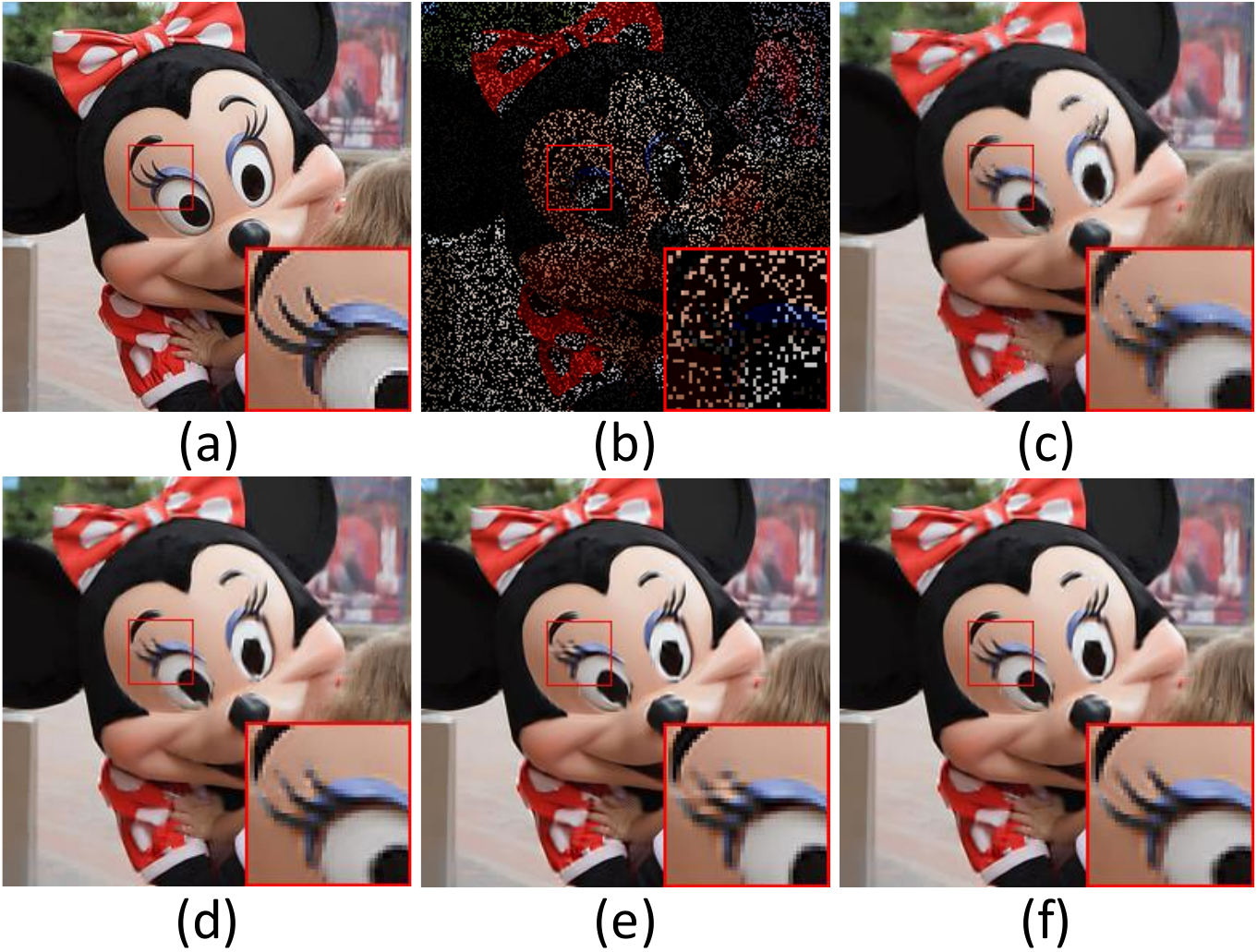}}
\end{minipage}
\vspace{-6mm}
	\caption{Inpainting performance comparison of $\emph{Mickey}$ based on the {\em designed adaptive dictionary learning} method. (a) Original image; (b) Degraded image with 80\% pixels missing; (c) $\ell_1$-norm (PSNR= 25.97dB); (d) $\ell_p$-norm (PSNR= 26.74dB); (e) $w\ell_1$-norm (PSNR= 26.66dB); (f) $w\ell_p$-norm (PSNR= \textbf{26.92dB}).}
	\label{fig:4}
\vspace{-4mm}
\end{figure}

\begin{figure}[!t]
	 \centering
\begin{minipage}[b]{1\linewidth}
{\includegraphics[width= 1\textwidth]{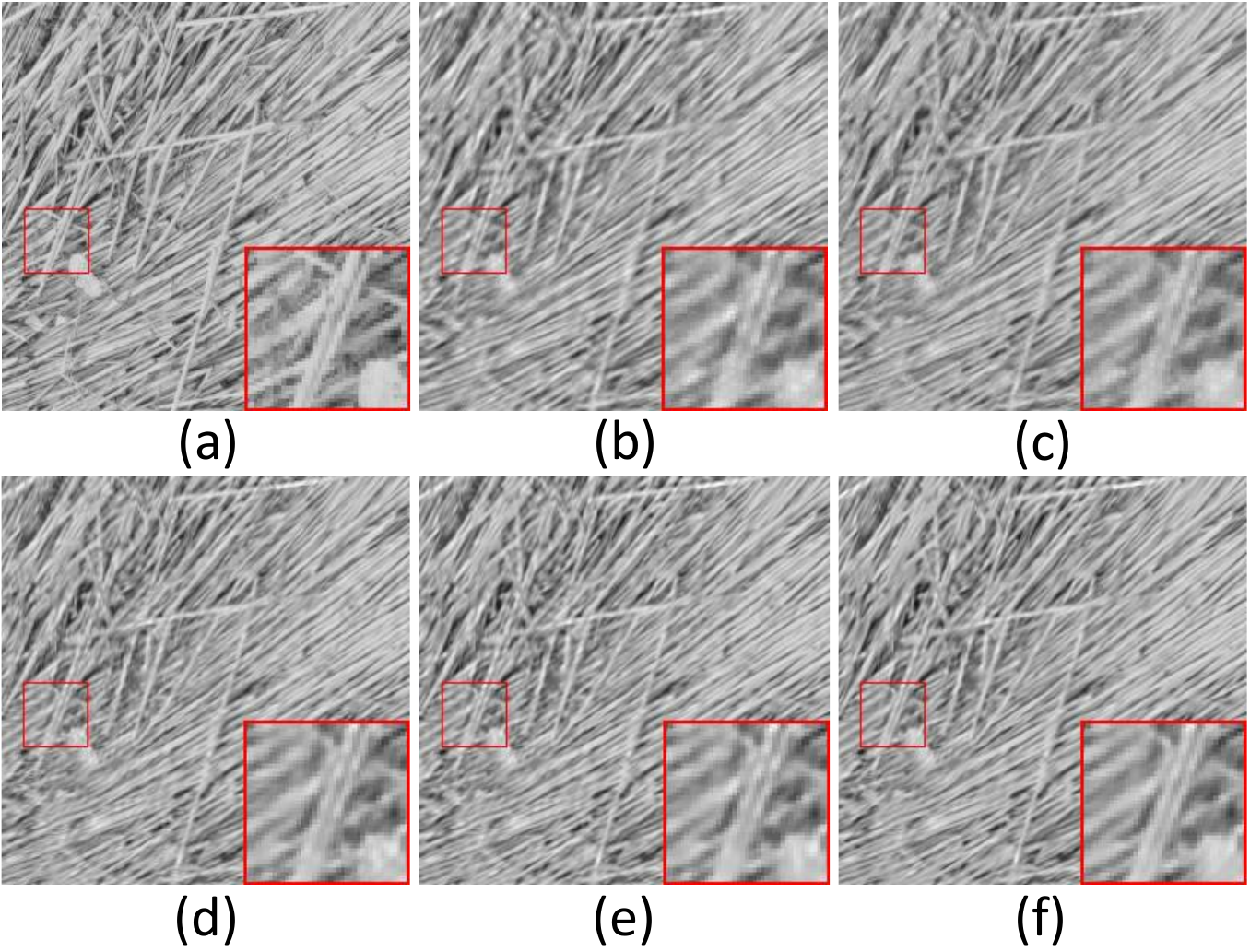}}
\end{minipage}
	\vspace{-6mm}
	\caption{CS recovery performance comparison of $\emph{Straw}$ with $0.2N$ measurements based on the {\em designed adaptive dictionary learning} method. (a) Original image; (b) Initial recovered image by \cite{40} (PSNR= 23.76dB); (c) $\ell_1$-norm (PSNR= 24.11dB); (d) $\ell_p$-norm (PSNR= 24.82dB); (e) $w\ell_1$-norm (PSNR= 24.82dB); (f) $w\ell_p$-norm (PSNR= \textbf{25.06dB}).}
	\label{fig:5}
\end{figure}

The PSNR results of image inpainting and image CS recovery are shown in Table~\ref{lab:2} and  Table~\ref{lab:3}, respectively. It can be seen that the  $w\ell_p$-norm minimization achieves better results than the three other norm  minimization methods in most cases in terms of PSNR. Fig.~\ref{fig:4} shows the image inpainting results of image $\emph{Mickey}$ with 80\% pixels missing. Fig.~\ref{fig:5} displays the image CS recovery results of image $\emph{Straw}$ with 0.2$N$ measurements. We can observe that  the $w\ell_p$-norm  minimization obtains better perceptual quality than the three other norm  minimization methods. Therefore, these experimental results are consistent with our previous analysis. This demonstrates the proposed benchmark is feasible.

\begin{table}[!t]
	\vspace{-2mm}
	\caption{PSNR ($\textnormal{d}$B) comparison of $\ell_1$-norm, $\ell_p$-norm, $w\ell_1$-norm and the $w\ell_p$-norm, based on the designed ADL method for image inpainting.}
\Huge
	\centering
	\resizebox{ 0.485\textwidth}{!}
	{\Huge
		\begin{tabular}{|c|c|c|c|c|c|c|c|c|c|c|c|}
			\hline
			\multirow{1}{*}{\textbf{Miss pixels}} &\multirow{1}{*}{\textbf{Methods}}&\multirow{1}{*}{\textbf{Mickey}}&\multirow{1}{*}{\textbf{Butterfly}}
			&\multirow{1}{*}{\textbf{Fence}}&\multirow{1}{*}{\textbf{Starfish}}
			&\multirow{1}{*}{\textbf{Nanna}}&\multirow{1}{*}{\textbf{Zebra}}
			&\multirow{1}{*}{\textbf{Fireman}}&\multirow{1}{*}{\textbf{Mural}}&\multirow{1}{*}{\textbf{Average}}\\
			\cline{2-11}
			\hline
			\multirow{4}{*}{80\%}
			& $\ell_1$-norm      & 25.97 & 25.61 & 28.90 & 26.98 & 25.46 & 22.33 & 25.39 & 25.29 & 25.74\\
			\cline{2-11}
			& $\ell_p$-norm       & 26.74 & 26.36 & 29.47 & 27.54 & 25.87 & 23.05 & 25.63 & 25.69 & 26.29\\
			\cline{2-11}
			& $w\ell_1$-norm      & 26.66 & 26.39 & 29.98 & 28.00 & 25.73 & 22.39 & 25.68 & 26.12 & 26.37\\
			\cline{2-11}
			& $w\ell_p$-norm      & \textbf{26.92} & \textbf{26.52} & \textbf{30.00} & \textbf{28.05} & \textbf{25.95} & \textbf{23.06} & \textbf{25.80} & \textbf{26.26} & \textbf{26.57}\\
			\hline
			\multirow{4}{*}{70\%}
			& $\ell_1$-norm      & 27.86 & 27.86 & 30.72 & 29.02 & 27.32 & 24.26 & 27.05 & 27.40 & 27.69\\
			\cline{2-11}
			& $\ell_p$-norm       & 29.04 & 29.10 & 31.54 & 30.25 & 28.31 & \textbf{25.19} & 27.69 & 28.37 & 28.69\\
			\cline{2-11}
			& $w\ell_1$-norm      & 29.16 & 29.21 & 31.83 & 30.54 & 28.07 & 24.82 & 27.81 & 28.53 & 28.74\\
			\cline{2-11}
			& $w\ell_p$-norm      & \textbf{29.29} & \textbf{29.28} & \textbf{31.85} & \textbf{30.56} & \textbf{28.39} & 25.13 & \textbf{27.84} & \textbf{28.62} & \textbf{28.87}\\
			\hline
			\multirow{4}{*}{60\%}
			& $\ell_1$-norm      & 29.61 & 29.82 & 32.29 & 30.78 & 29.02 & 25.93 & 28.53 & 28.93 & 29.36\\
			\cline{2-11}
			& $\ell_p$-norm       & 29.85 & 30.16 & 32.51 & 31.14 & 29.22 & 26.13 & 28.69 & 29.16 & 29.61\\
			\cline{2-11}
			& $w\ell_1$-norm      & 31.44 & 31.40 & 33.65 & 32.93 & 30.42 & 27.04 & 29.74 & 30.27 & 30.86\\
			\cline{2-11}
			& $w\ell_p$-norm      & \textbf{31.46} & \textbf{31.54} & \textbf{33.67} & \textbf{33.02} & \textbf{30.56} & \textbf{27.21} & \textbf{29.77} & \textbf{30.35} & \textbf{30.95}\\
			\hline
			\multirow{4}{*}{50\%}
			& $\ell_1$-norm      & 31.62 & 31.46 & 33.78 & 32.62 & 30.68 & 27.65 & 30.13 & 30.48 & 31.05\\
			\cline{2-11}
			& $\ell_p$-norm       & 31.88 & 31.78 & 33.97 & 32.99 & 30.89 & 27.86 & 30.28 & 30.70 & 31.29\\
			\cline{2-11}
			& $w\ell_1$-norm      & 33.98 & 33.16 & \textbf{35.30} & 34.99 & 32.38 & 29.12 & 31.31 & 31.88 & 32.76\\
			\cline{2-11}
			& $w\ell_p$-norm      & \textbf{34.01} & \textbf{33.26} & 35.25 & \textbf{35.05} & \textbf{32.53} & \textbf{29.26} & \textbf{31.32} & \textbf{31.91} & \textbf{32.82}\\
			\hline
	\end{tabular}}
	\label{lab:2}
	\vspace{-2mm}
\end{table}

\begin{table}[!t]
	\caption{PSNR ($\textnormal{d}$B) comparison of $\ell_1$-norm, $\ell_p$-norm, $w\ell_1$-norm and the $w\ell_p$-norm, based on the designed ADL method for image CS Recovery.}
\Huge
	\centering  
		\resizebox{0.485\textwidth}{!}
	{
	\begin{tabular}{|c|c|c|c|c|c|c|c|c|c|c|c|}
		\hline
		\multirow{1}{*}{\textbf{Ratio}} &\multirow{1}{*}{\textbf{Methods}}&\multirow{1}{*}{\textbf{Barbara}}&\multirow{1}{*}{\textbf{Bridge}}
		&\multirow{1}{*}{\textbf{Elaine}}&\multirow{1}{*}{\textbf{Fence}}
		&\multirow{1}{*}{\textbf{House}}&\multirow{1}{*}{\textbf{Lena}}
		&\multirow{1}{*}{\textbf{Peppers}}&\multirow{1}{*}{\textbf{Straw}}&\multirow{1}{*}{\textbf{Average}}\\
		\cline{2-11}
		\hline
		\multirow{4}{*}{0.2}
		& $\ell_1$-norm      & 32.24 & 25.03 & 34.59 & 29.31 & 36.01 & 30.77 & 30.00 & 24.11 & 30.26\\
		\cline{2-11}
		& $\ell_p$-norm       & 34.31 & 25.13 & 35.75 & 29.99 & \textbf{37.15} & 31.50 & 30.79 & 24.82 & 31.18\\
		\cline{2-11}
		& $w\ell_1$-norm      & 34.53 & 25.04 & \textbf{36.22} & 30.19 & {37.07} & 31.49 & 31.22 & 24.82 & 31.32\\
		\cline{2-11}
		& $w\ell_p$-norm      & \textbf{34.55} & \textbf{25.28} & 36.00 & \textbf{30.38} & 36.92 & \textbf{31.62} & \textbf{31.32} & \textbf{25.06} & \textbf{31.39}\\
		\hline
		\multirow{4}{*}{0.3}
		& $\ell_1$-norm      & 34.49 & 26.49 & 36.76 & 31.16 & 37.94 & 32.97 & 31.93 & 26.11 & 32.23\\
		\cline{2-11}
		& $\ell_p$-norm       & 34.86 & 26.59 & 37.07 & 31.42 & 38.29 & 33.15 & 32.22 & 26.23 & 32.48\\
		\cline{2-11}
		& $w\ell_1$-norm      & 37.10 & \textbf{27.25} & 38.26 & 32.50 & 39.07 & 34.26 & \textbf{33.39} & 27.84 & 33.71\\
		\cline{2-11}
		& $w\ell_p$-norm      & \textbf{37.23} & 27.22 & \textbf{38.30} & \textbf{32.53} & \textbf{39.23} & \textbf{34.29} & 33.32 & \textbf{27.89} & \textbf{33.75}\\
		\hline
		\multirow{4}{*}{0.4}
		& $\ell_1$-norm      & 36.79 & 27.94 & 38.58 & 32.84 & 39.70 & 34.76 & 33.63 & 27.95 & 34.03\\
		\cline{2-11}
		& $\ell_p$-norm       & 37.15 & 28.06 & 38.87 & 33.09 & 39.98 & 34.95 & 33.91 & 28.10 & 34.26\\
		\cline{2-11}
		& $w\ell_1$-norm      & 39.04 & \textbf{28.90} & 40.03 & \textbf{34.50} & 40.82 & 36.58 & \textbf{35.10} & \textbf{30.30} & 35.66\\
		\cline{2-11}
		& $w\ell_p$-norm      & \textbf{39.13} & 28.85 & \textbf{40.05} & 34.42 & \textbf{40.93} & \textbf{36.66} & 35.00 & 30.28 & \textbf{35.67}\\
		\hline
		\multirow{4}{*}{0.5}
		& $\ell_1$-norm      & 38.80 & 29.38 & 40.26 & 34.50 & 41.27 & 36.56 & 35.18 & 29.88 & 35.73\\
		\cline{2-11}
		& $\ell_p$-norm       & 39.19 & 29.51 & 40.54 & 34.75 & 41.52 & 36.78 & 35.43 & 30.07 & 35.97\\
		\cline{2-11}
		& $w\ell_1$-norm      & 40.84 & {30.51} & 41.52 & \textbf{36.29} & 42.25 & 38.99 & \textbf{36.56} & \textbf{32.49} & 37.43\\
		\cline{2-11}
		& $w\ell_p$-norm      & \textbf{40.94} & \textbf{30.52} & \textbf{41.63} & {36.24} & \textbf{42.38} & \textbf{39.09} & {36.53} & {32.46} & \textbf{37.47}\\
		\hline
	\end{tabular}}
	\label{lab:3}
\end{table}

\begin{table}[!t]
	\vspace{-2mm}
\caption{Detailed setting of $\rho, \lambda$ for image inpainting and image CS recovery, with the  graph-based dictionary \cite{38} and  PCA dictionary \cite{39}, respectively.}
\centering  
	\scriptsize
\resizebox{0.48\textwidth}{!}
{
\begin{tabular}{|c||c|c||c|c||c||c|c|c|}
\hline
\multicolumn{7}{|c|}{Image Inpainting}\\
\hline
  \multicolumn{1}{|c||}{Graph}&\multicolumn{2}{|c||}{$\ell_1$-norm}&\multicolumn{2}{|c||}{$\ell_p$-norm}&\multicolumn{1}{|c||}{$w\ell_1$-norm}&\multicolumn{1}{|c|}{$w\ell_p$-norm}\\
\hline
\multirow{1}{*}{{{Parameters}}}&\multirow{1}{*}{{{$\rho$}}}&\multirow{1}{*}{{{$\lambda$}}}&\multirow{1}{*}{{{$\rho$}}}
&{{{$\lambda$}}}&\multirow{1}{*}{{{$\rho$}}}&\multirow{1}{*}{{{$\rho$}}}\\
\hline
 \multirow{1}{*}{80\%}     &    0.008    &  1e-5      & 0.003      & 7e-5   & 0.15      & 0.06    \\
\hline
 \multirow{1}{*}{70\%}     &    0.003    &  1e-5      & 0.003      & 7e-5   & 0.1       & 0.05    \\
\hline
 \multirow{1}{*}{60\%}     &    0.006    &  7e-5      & 0.003      & 5e-5   & 0.07      & 0.09    \\
\hline
 \multirow{1}{*}{50\%}     &    0.003    &  7e-5      & 0.003      & 9e-5   & 0.05      & 0.05    \\
\hline
\multicolumn{7}{|c|}{Image CS Recovery}\\
\hline
  \multicolumn{1}{|c||}{PCA}&\multicolumn{2}{|c||}{$\ell_1$-norm}&\multicolumn{2}{|c||}{$\ell_p$-norm}&\multicolumn{1}{|c||}{$w\ell_1$-norm}&\multicolumn{1}{|c|}{$w\ell_p$-norm}\\
\hline
\multirow{1}{*}{{{Parameters}}}&\multirow{1}{*}{{{$\rho$}}}&\multirow{1}{*}{{{$\lambda$}}}&\multirow{1}{*}{{{$\rho$}}}
&{{{$\lambda$}}}&\multirow{1}{*}{{{$\rho$}}}&\multirow{1}{*}{{{$\rho$}}}\\
\hline
 \multirow{1}{*}{0.2$N$}     &    0.03     &  1e-6      & 0.003      & 7e-5   & 0.09     & 0.07    \\
\hline
 \multirow{1}{*}{0.3$N$}     &    0.008    &  1e-6      & 0.006      & 7e-5   & 0.07     & 0.09    \\
\hline
 \multirow{1}{*}{0.4$N$}     &    0.008    &  1e-6      & 0.008      & 1e-5   & 0.05     & 0.05    \\
\hline
 \multirow{1}{*}{0.5$N$}     &    0.008    &  1e-6      & 0.008      & 1e-5   & 0.05     & 0.05    \\
\hline
\end{tabular}
\label{lab:4}
}
\vspace{-4mm}
\end{table}

\begin{figure}[!t]

\begin{minipage}[b]{1\linewidth}
{\includegraphics[width= 1\textwidth]{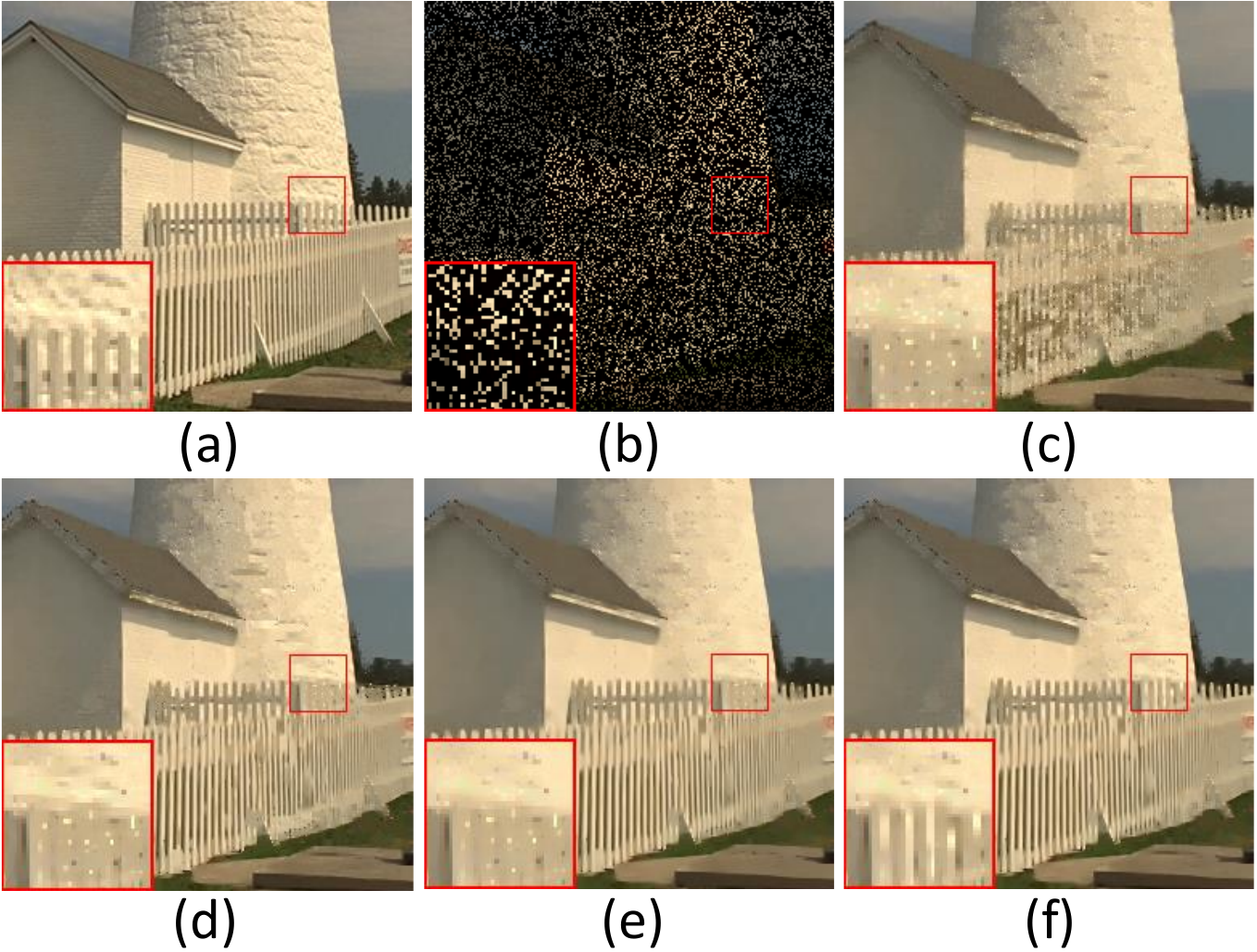}}
\end{minipage}
\vspace{-6mm}
\caption{Inpainting performance comparison on the image $\emph{Fence}$ based on the {\em graph-based dictionary learning} method \cite{38}. (a) Original image; (b) Degraded image with 80\% pixels missing sample; (c) $\ell_1$-norm (PSNR= 22.92dB); (d) $\ell_p$-norm (PSNR= 25.83dB); (e) $w\ell_1$-norm (PSNR= 26.67dB); (f) $w\ell_p$-norm (PSNR= \textbf{26.99dB}).}
\label{fig:6}
\vspace{-2mm}
\end{figure}

\begin{figure}[!t]
\begin{minipage}[b]{1\linewidth}
{\includegraphics[width= 1\textwidth]{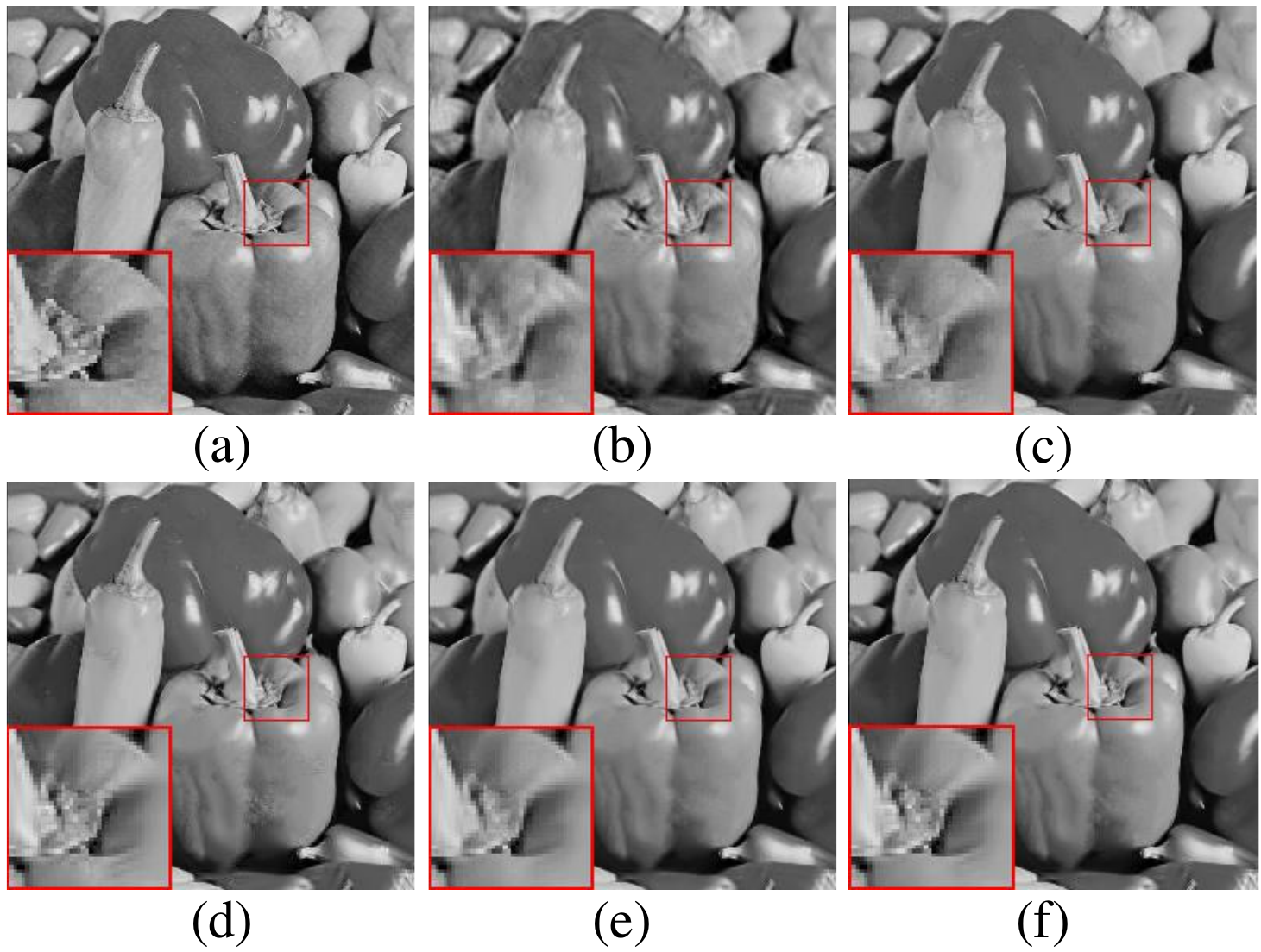}}
\end{minipage}
\vspace{-6mm}
\caption{CS recovery performance comparison with $0.2N$ measurements on the image $\emph{Peppers}$ based on the {\em PCA dictionary learning} method \cite{39}. (a) Original image; (b) Initial image by \cite{40} (PSNR= 28.61dB); (c) $\ell_1$-norm (PSNR= 30.08dB); (d) $\ell_p$-norm (PSNR= 30.87dB); (e) $w\ell_1$-norm (PSNR= 31.10dB); (f) $w\ell_p$-norm (PSNR= \textbf{31.31dB}).}
\vspace{-4mm}
\label{fig:7}
\end{figure}

{Next, in order to further prove the universality and advantages of the proposed scheme, instead of using the designed ADL method, we exploit another two generalized dictionary learning methods to verify the feasibility of the proposed benchmark, \ie, graph-based dictionary learning method \cite{38} and PCA dictionary learning method \cite{39} for image inpainting and image CS recovery, respectively. Similar to the designed ADL method, we learn the graph-based dictionary and the PCA dictionary from each patch group of the degraded image. All the parameters remain the same as specified in the subsection~\ref{5.1} except for $\rho$ and $\lambda$, which are now shown in Table~\ref{lab:4}. The PSNR comparison results for image inpainting of four competing methods are shown in Table~\ref{lab:5}.
It can be seen that the $w\ell_p$-norm  minimization consistently outperforms the three other  norm  minimization  methods for all test images (the only exception is the image $\emph{Zebra}$ for which the $w\ell_1$-norm  minimization  is slightly higher than  the $w\ell_p$-norm  minimization  in the scene of 80\% and 70\% pixels missing).
The PSNR comparison results for image CS recovery are shown in Table~\ref{lab:6}, and we can observe that the  $w\ell_p$-norm  minimization outperforms the three other norm  minimization methods in most cases. Fig.~\ref{fig:6} shows the visual comparison of image $\emph{Fence}$ with 80\% pixels missing for image inpainting based on the graph-based dictionary learning method. The visual comparison of image $\emph{Peppers}$ with 0.2$N$ measurements for image CS recovery based on the PCA dictionary learning method is shown in Fig.~\ref{fig:7}. It can be observed that the  $w\ell_p$-norm  minimization  achieves better visual quality than the  three other norm  minimization methods. This again verifies the feasibility of the proposed benchmark.}

\begin{table}[!t]
\vspace{-4mm}
	\caption{PSNR ($\textnormal{d}$B) comparison of $\ell_1$-norm, $\ell_p$-norm, $w\ell_1$-norm and the $w\ell_p$-norm, based on the graph-based dictionary learning method \cite{38} for image inpainting.}
	\centering  
\Huge
	\resizebox{0.485\textwidth}{!}
	{
		\begin{tabular}{|c|c|c|c|c|c|c|c|c|c|c|c|}
			\hline
			\multirow{1}{*}{\textbf{Miss pixels}} &\multirow{1}{*}{\textbf{Methods}}&\multirow{1}{*}{\textbf{Mickey}}&\multirow{1}{*}{\textbf{Butterfly}}
			&\multirow{1}{*}{\textbf{Fence}}&\multirow{1}{*}{\textbf{Starfish}}
			&\multirow{1}{*}{\textbf{Nanna}}&\multirow{1}{*}{\textbf{Zebra}}
			&\multirow{1}{*}{\textbf{Fireman}}&\multirow{1}{*}{\textbf{Mural}}&\multirow{1}{*}{\textbf{Average}}\\
			\cline{2-11}
			\hline
			\multirow{4}{*}{80\%}
			& $\ell_1$-norm      & 24.63 & 23.62 & 22.92 & 26.27 & 24.52 & 20.45 & 24.59 & 23.57 & 23.82\\
			\cline{2-11}
			& $\ell_p$-norm       & 25.20 & 24.51 & 25.83 & 26.41 & 24.79 & 20.79 & 24.88 & 24.31 & 24.59\\
			\cline{2-11}
			& $w\ell_1$-norm      & 25.30 & 24.63 & 26.67 & 26.20 & 24.75 & \textbf{20.98} & 24.88 & 24.44 & 24.73\\
			\cline{2-11}
			& $w\ell_p$-norm      & \textbf{25.40} & \textbf{24.74} & \textbf{26.99} & \textbf{26.43} & \textbf{24.92} & 20.97 & \textbf{25.00} & \textbf{24.60} & \textbf{24.88}\\
			\hline
			\multirow{4}{*}{70\%}
			& $\ell_1$-norm      & 26.31 & 25.75 & 25.06 & 28.07 & 26.21 & 21.84 & 26.08 & 25.33 & 25.58\\
			\cline{2-11}
			& $\ell_p$-norm       & 27.21 & 26.88 & 28.27 & 28.33 & 26.58 & 22.43 & 26.50 & 26.43 & 26.58\\
			\cline{2-11}
			& $w\ell_1$-norm      & 27.37 & 27.00 & 28.89 & 27.92 & 26.58 & \textbf{22.80} & 26.55 & 26.65 & 26.72\\
			\cline{2-11}
			& $w\ell_p$-norm      & \textbf{27.49} & \textbf{27.19} & \textbf{29.12} & \textbf{28.50} & \textbf{26.81} & 22.77 & \textbf{26.77} & \textbf{26.84} & \textbf{26.93}\\
			\hline
			\multirow{4}{*}{60\%}
			& $\ell_1$-norm      & 27.58 & 27.50 & 27.48 & 29.46 & 27.61 & 23.37 & 27.38 & 26.80 & 27.15\\
			\cline{2-11}
			& $\ell_p$-norm       & 27.72 & 27.66 & 28.09 & 29.47 & 27.65 & 23.46 & 27.41 & 26.93 & 27.30\\
			\cline{2-11}
			& $w\ell_1$-norm      & 28.79 & 28.84 & 30.74 & 29.68 & 28.22 & 24.72 & 28.05 & 28.39 & 28.43\\
			\cline{2-11}
			& $w\ell_p$-norm      & \textbf{28.80} & \textbf{28.91} & \textbf{30.79} & \textbf{29.81} & \textbf{28.29} & \textbf{24.75} & \textbf{28.11} & \textbf{28.45} & \textbf{28.49}\\
			\hline
			\multirow{4}{*}{50\%}
			& $\ell_1$-norm      & 29.14 & 29.24 & 29.73 & 30.93 & 29.13 & 24.84 & 28.78 & 28.37 & 28.77\\
			\cline{2-11}
			& $\ell_p$-norm       & 29.31 & 29.40 & 30.24 & 30.86 & 29.12 & 24.98 & 28.80 & 28.50 & 28.90\\
			\cline{2-11}
			& $w\ell_1$-norm      & 30.68 & 30.62 & 32.65 & 31.51 & 29.76 & 26.61 & 29.64 & 30.04 & 30.19\\
			\cline{2-11}
			& $w\ell_p$-norm      & \textbf{30.71} & \textbf{30.69} & \textbf{32.71} & \textbf{31.56} & \textbf{29.82} & \textbf{26.62} & \textbf{29.70} & \textbf{30.08} & \textbf{30.24}\\
			\hline
	\end{tabular}}
	\label{lab:5}
\end{table}

\begin{table}[!t]
	\caption{PSNR ($\textnormal{d}$B) comparison of $\ell_1$-norm, $\ell_p$-norm, $w\ell_1$-norm and the $w\ell_p$-norm, based on the PCA dictionary learning method \cite{39} for image CS Recovery.}
\Huge
	\centering  
	\resizebox{0.485\textwidth}{!}
	{
		\begin{tabular}{|c|c|c|c|c|c|c|c|c|c|c|c|}
			\hline
			\multirow{1}{*}{\textbf{Ratio}} &\multirow{1}{*}{\textbf{Methods}}&\multirow{1}{*}{\textbf{Barbara}}&\multirow{1}{*}{\textbf{Bridge}}
			&\multirow{1}{*}{\textbf{Elaine}}&\multirow{1}{*}{\textbf{Fence}}
			&\multirow{1}{*}{\textbf{House}}&\multirow{1}{*}{\textbf{Lena}}
			&\multirow{1}{*}{\textbf{Peppers}}&\multirow{1}{*}{\textbf{Straw}}&\multirow{1}{*}{\textbf{Average}}\\
			\cline{2-11}
			\hline
			\multirow{4}{*}{0.2}
			& $\ell_1$-norm      & 32.13 & 25.05 & 34.60 & 29.12 & 35.92 & 30.78 & 30.08 & 24.10 & 30.22\\
			\cline{2-11}
			& $\ell_p$-norm       & 34.07 & 25.24 & 35.74 & 29.87 & \textbf{36.76} & 31.46 & 30.87 & 24.74 & 31.09\\
			\cline{2-11}
			& $w\ell_1$-norm      & 32.89 & 25.25 & 33.32 & 29.87 & 36.23 & 31.23 & 31.10 & 24.50 & 30.55\\
			\cline{2-11}
			& $w\ell_p$-norm      & \textbf{34.09} & \textbf{25.29} & \textbf{35.74} & \textbf{30.14} & 36.75 & \textbf{31.48} & \textbf{31.31} & \textbf{24.89} & \textbf{31.21}\\
			\hline
			\multirow{4}{*}{0.3}
			& $\ell_1$-norm      & 34.38 & 26.53 & 36.80 & 31.08 & 38.05 & 33.03 & 32.10 & 26.09 & 32.26\\
			\cline{2-11}
			& $\ell_p$-norm       & 34.50 & 26.59 & 36.84 & 31.19 & 38.04 & 33.09 & 32.24 & 26.18 & 32.33\\
			\cline{2-11}
			& $w\ell_1$-norm      & 35.53 & 26.95 & 34.94 & 31.73 & 38.04 & 33.69 & 32.99 & 27.05 & 32.61\\
			\cline{2-11}
			& $w\ell_p$-norm      & \textbf{35.69} & \textbf{27.04} & \textbf{37.25} & \textbf{31.84} & \textbf{38.07} & \textbf{33.80} & \textbf{33.06} & \textbf{27.28} & \textbf{33.01}\\
			\hline
			\multirow{4}{*}{0.4}
			& $\ell_1$-norm      & 36.69 & 28.01 & 38.65 & 32.84 & 39.80 & 34.83 & 33.85 & 27.96 & 34.08\\
			\cline{2-11}
			& $\ell_p$-norm       & 37.03 & 28.13 & 38.92 & 33.09 & \textbf{40.07} & 35.04 & 34.10 & 28.10 & 34.31\\
			\cline{2-11}
			& $w\ell_1$-norm      & 37.63 & 28.39 & 39.03 & 33.43 & 39.75 & 35.46 & 34.59 & 29.07 & 34.67\\
			\cline{2-11}
			& $w\ell_p$-norm      & \textbf{37.74} & \textbf{28.45} & \textbf{39.11} & \textbf{33.51} & 39.84 & \textbf{35.57} & \textbf{34.65} & \textbf{29.21} & \textbf{34.76}\\
			\hline
			\multirow{4}{*}{0.5}
			& $\ell_1$-norm      & 38.72 & 29.46 & 40.32 & 34.55 & 41.36 & 36.64 & 35.45 & 29.89 & 35.80\\
			\cline{2-11}
			& $\ell_p$-norm       & 39.09 & 29.60 & 40.59 & 34.80 & \textbf{41.59} & 36.88 & 35.66 & 30.07 & 36.03\\
			\cline{2-11}
			& $w\ell_1$-norm      & 39.53 & 29.95 & 40.59 & 35.13 & 41.25 & 37.45 & 36.09 & 31.16 & 36.39\\
			\cline{2-11}
			& $w\ell_p$-norm      & \textbf{39.64} & \textbf{30.00} & \textbf{40.67} & \textbf{35.21} & 41.33 & \textbf{37.56} & \textbf{36.14} & \textbf{31.30} & \textbf{36.48}\\
			\hline
	\end{tabular}}
	\label{lab:6}
	\vspace{-2mm}
\end{table}

We also notice that for various dictionary learning methods, the results of the $w\ell_p$-norm  minimization are usually a little bit ($\sim$0.1dB) better than those of the $w\ell_1$-norm  minimization, which is consistent with the observation in  \cite{27}. 
Comparing Tables~\ref{lab:2}-\ref{lab:3} with Tables~\ref{lab:5}-\ref{lab:6}, we can see that the designed ADL method can provide better performance than the graph-based dictionary learning and PCA dictionary learning methods. This demonstrates the superiority of the designed adaptive dictionary learning method, while further illustrating the feasibility of the proposed benchmark in sparse coding.   

\subsection{Comparison with Other Leading Algorithms}

We now validate the performance of the proposed scheme, \ie, group-based sparse coding with non-convex $w\ell_p$-norm (GSC-$w\ell_p$) minimization for image inpainting and image CS recovery with the designed adaptive dictionary learning method, through comparing it with recent state-of-the-art methods\footnote{We would like to thank the authors of \cite{40,41,42,43,44,45,46,47,48,49,50,64,65,66,70,71} for kindly providing their software or codes.}.

\begin{table*}[!t]
	\vspace{-4mm}
\caption{PSNR ($\textnormal{d}$B) comparison of SALSA \cite{41}, BPFA \cite{42}, IPPO \cite{43}, JSM \cite{44}, Aloha \cite{45}, BKSVD \cite{64}, NGS \cite{65}, IRCNN \cite{70}, IDBP \cite{71} and GSC-$w\ell_p$ for image inpainting.}
\centering
	\resizebox{ 0.80\textwidth}{!}
	{
		\begin{tabular}{|c|c|c|c|c|c|c|c|c|c|c|c|}
			\hline
			\multirow{1}{*}{\textbf{Miss pixels}} &\multirow{1}{*}{\textbf{Methods}}&\multirow{1}{*}{\textbf{Mickey}}&\multirow{1}{*}{\textbf{Butterfly}}
			&\multirow{1}{*}{\textbf{Fence}}&\multirow{1}{*}{\textbf{Starfish}}
			&\multirow{1}{*}{\textbf{Nanna}}&\multirow{1}{*}{\textbf{Zebra}}
			&\multirow{1}{*}{\textbf{Fireman}}&\multirow{1}{*}{\textbf{Mural}}&\multirow{1}{*}{\textbf{Average}}\\
			\cline{2-11}
			\hline
			\multirow{10}{*}{80\%}
			&    SALSA       & 24.46 & 22.85 & 21.80 & 25.70 & 24.12 & 19.68 & 24.38 & 23.15 & 23.27\\
			\cline{2-11}
			&     BPFA        & 24.53 & 24.04 & 26.24 & 26.79 & 24.71 & 20.90 & 24.88 & 24.13 & 24.53\\
			\cline{2-11}
			&     IPPO       & 26.33 & 25.13 & 27.98 & 26.30 & 25.60 & 22.71 & 25.56 & 25.66 & 25.66\\
			\cline{2-11}
			&     JSM        & 26.09 & 25.57 &28.59  & 27.07 & 25.33 & 21.88 & 25.31 & 25.40 & 25.65\\
			\cline{2-11}
			&     Aloha       & 25.33 & 24.88 &28.88  & 26.33 & 25.54 & 22.72 & 25.03 & 25.23 & 25.49\\

			\cline{2-11}
			&     BKSVD       & 23.72 & 22.00 & 24.20  & 25.36 & 23.97 & 19.37 & 23.79 & 23.02 & 23.18\\
			\cline{2-11}
			&     NGS         & 24.50 & 23.85 & 25.26 & 26.17 & 24.58 & 20.49 & 24.54 & 23.78 & 24.15\\
			\cline{2-11}
			&     IRCNN       &26.45 &	25.34 &	27.76 &	27.01 &	25.76 &	22.43 &	25.78 &	25.75 &	25.78\\
			\cline{2-11}
			&     IDBP        &	25.40 &	25.24 &	25.03 &	26.88 &	25.51 &	20.97 &	25.42 &	25.26 &	24.96\\
            \cline{2-11}
			& GSC-$w\ell_p$     & \textbf{26.92} & \textbf{26.52} & \textbf{30.00}  & \textbf{28.05} & \textbf{25.95} & \textbf{23.06} & \textbf{25.80} & \textbf{26.26} & \textbf{26.57}\\
			\hline

			\multirow{10}{*}{70\%}
			&    SALSA       & 25.98 & 25.06 & 23.57 & 27.55 & 25.44 & 21.41 & 25.82 & 25.00 & 24.98\\
			\cline{2-11}
			&     BPFA        & 26.16 & 26.68 & 28.87 & 28.93 & 26.62 & 22.78 & 26.55 & 26.46 & 26.63\\
			\cline{2-11}
			&     IPPO        & 28.59 & 27.68 & 30.08 & 28.91 & 27.44 & 24.76 & 27.44 & 27.92 & 27.85\\
			\cline{2-11}
			&     JSM         & 28.25 & 27.97 & 30.46 & 29.36 & 27.34 & 23.95 & 27.16 & 27.59 & 27.76\\
			\cline{2-11}
			&     Aloha       & 27.11 & 27.29 & 30.57 & 28.22 & 27.43 & 24.55 & 26.52 & 27.33 & 27.38\\
			\cline{2-11}
			&     BKSVD       & 26.17 & 25.00 & 28.35  & 27.79 & 26.18 & 23.06 & 25.85 & 25.57 & 26.00\\

			\cline{2-11}
			&     NGS        & 26.68 & 26.36 & 27.32 & 28.35 & 26.35 & 22.71 & 26.29 & 26.06 & 26.27\\
			\cline{2-11}
			&     IRCNN       &	\textbf{29.65} &28.33 &	30.53 &	29.78 &	28.27 &	25.04 &	\textbf{27.87} &\textbf{28.72} &28.53\\
			\cline{2-11}
			&     IDBP        &	28.69 &	28.29 &	29.24 &	29.28 &	27.24 &	23.60 &	27.63 &	27.68 &	27.71\\
			\cline{2-11}
			& GSC-$w\ell_p$     & {29.29} & \textbf{29.28} & \textbf{31.85}  & \textbf{30.56} & \textbf{28.39} & \textbf{25.13} & {27.84} & {28.61} & \textbf{28.87}\\
			\hline

			\multirow{10}{*}{60\%}
			&    SALSA       & 27.41 & 26.79 & 25.45 & 29.09 & 26.94 & 22.80 & 27.15 & 26.66 & 26.54\\
			\cline{2-11}
			&     BPFA        & 27.83 & 28.88 & 30.79 & 30.98 & 28.63 & 24.53 & 28.23 & 28.30 & 28.52\\
			\cline{2-11}
			&     IPPO        & 30.76 & 29.85 & 32.14 & 31.09 & 29.41 & 26.79 & 29.13 & 29.57 & 29.84\\
			\cline{2-11}
			&     JSM        & 29.85 & 29.83 & 32.23 & 31.40 & 29.09 & 25.90 & 28.79 & 29.24 & 29.54\\
			\cline{2-11}
			&     Aloha       & 28.59 & 29.16 & 32.33 & 30.19 & 29.51 & 26.24 & 28.24 & 28.92 & 29.15\\
			\cline{2-11}
			&     BKSVD       & 28.53 & 27.70 & 30.72  & 29.99 & 28.35 & 25.27 & 27.97 & 27.90 & 28.30\\
			\cline{2-11}
			&     NGS        & 28.09 & 28.37 & 30.11 & 30.26 & 28.06 & 24.39 & 27.67 & 27.99 & 28.12\\
			\cline{2-11}
			&     IRCNN       &	\textbf{31.81}	&30.41&	32.45 &	32.28 &	30.48 &	27.03 &	\textbf{29.85} &\textbf{30.57} &30.61\\
			\cline{2-11}
			&     IDBP       &	31.18 &	30.03 &	31.25 &	31.79 &	29.51 &	25.97 &	29.28 &	29.79 &	29.85\\
\cline{2-11}
			& GSC-$w\ell_p$     & 31.46 & \textbf{31.54} & \textbf{33.67}  & \textbf{33.02} & \textbf{30.56} & \textbf{27.21} & {29.77} & {30.35} & \textbf{30.95}\\
			\hline

			\multirow{10}{*}{50\%}
			&    SALSA       & 28.98 & 28.52 & 27.25 & 30.90 & 28.53 & 24.42 & 28.54 & 28.20 & 28.17\\
			\cline{2-11}
			&     BPFA        & 29.43 & 30.98 & 32.82 & 33.13 & 30.68 & 26.37 & 30.12 & 30.46 & 30.50\\
			\cline{2-11}
			&     IPPO        & 32.74 & 31.69 & 33.95 & 33.10 & 31.17 & 28.42 & 30.82 & 31.11 & 31.63\\
			\cline{2-11}
			&     JSM         & 31.96 & 31.47 & 33.75 & 33.24 & 30.75 & 27.77 & 30.37 & 30.89 & 31.27\\
			\cline{2-11}
			&     Aloha       & 30.33 & 30.78 & 33.79 & 31.85 & 31.24 & 27.67 & 29.88 & 30.28 & 30.73\\
			\cline{2-11}
			&     BKSVD       & 29.95 & 29.64 & 32.44  & 31.99 & 30.15 & 26.97 & 29.47 & 29.59 & 30.02\\
			\cline{2-11}
			&     NGS        & 29.75 & 30.28 & 32.00 & 32.10 & 29.71 & 26.03 & 29.22 & 29.88 & 29.87\\
			\cline{2-11}
			&     IRCNN       &	\textbf{34.22} &32.47 &	34.30&34.47 &32.41&	29.17 &	\textbf{31.45}&	\textbf{32.19}&	32.59\\
			\cline{2-11}
			&     IDBP        &	33.14 &	32.44 &	33.24 &	33.81 &	31.35 &	28.65 &	30.97 &	31.35 &	31.87\\
			\cline{2-11}
			& GSC-$w\ell_p$     & {34.00} & \textbf{33.26} & \textbf{35.25}  & \textbf{35.05} & \textbf{32.53} & \textbf{29.26} & {31.32} & {31.91} & \textbf{32.82}\\
			\hline
	\end{tabular}}
	\label{lab:7}
	\vspace{-2mm}
\end{table*}

\begin{figure}[!t]
\begin{minipage}[b]{1\linewidth}
{\includegraphics[width= 1\textwidth]{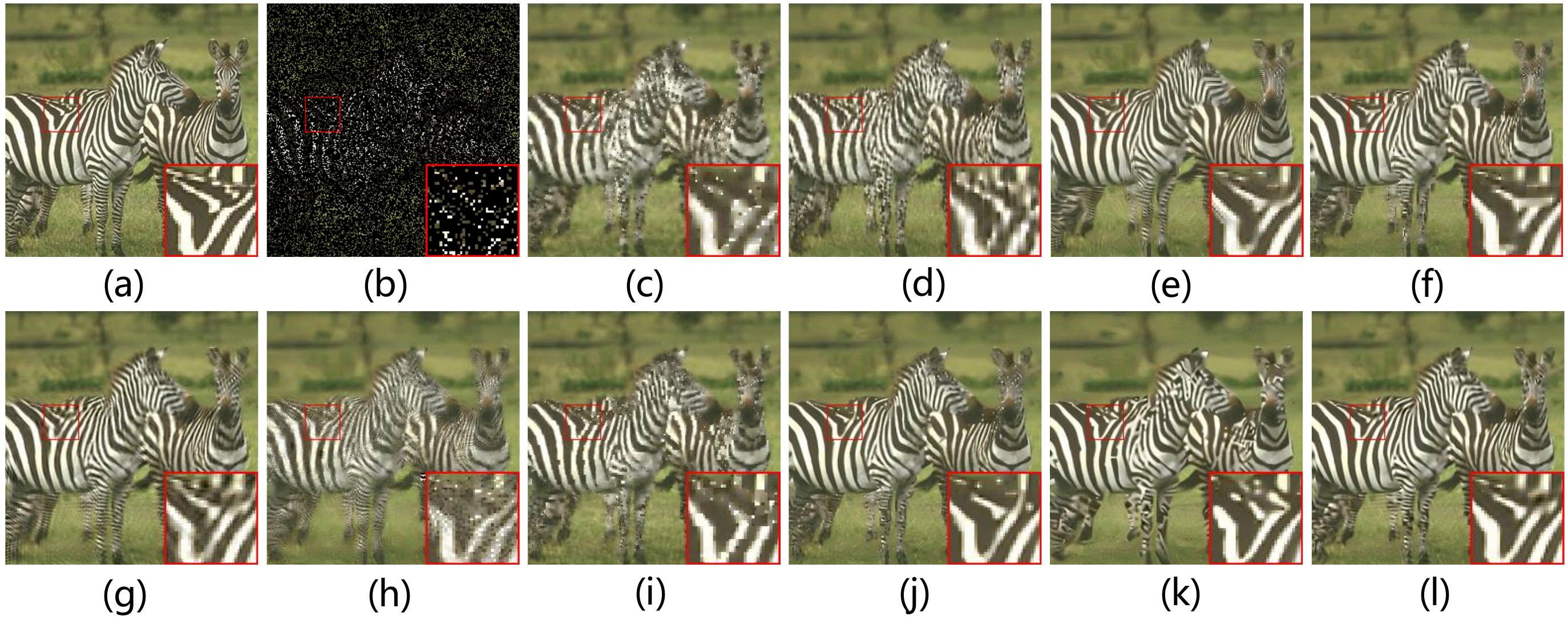}}
\end{minipage}
	\vspace{-6mm}
	\caption{Inpainting performance comparison on the image $\emph{Zebra}$. (a) Original image; (b) Degraded image with 80\% pixels missing sample; (c) SALSA \cite{41} (PSNR = 19.68dB); (d) BPFA \cite{42} (PSNR = 20.90dB); (e) IPPO \cite{43} (PSNR = 22.71dB); (f) JSM \cite{44} (PSNR = 21.88dB); (g) Aloha \cite{45} (PSNR = 22.72dB);  (h) BKSVD \cite{64} (PSNR = 19.37dB);   (i) NGS \cite{65} (PSNR = 20.49dB);  (j) IRCNN \cite{70} (PSNR = 22.43dB);   (k) IDBP \cite{71} (PSNR = 20.97dB);  (l) GSC-$w\ell_p$ (PSNR = \textbf{23.06dB}).}
	\label{fig:8}
	\vspace{-4mm}
\end{figure}

\begin{figure}[!t]
\begin{minipage}[b]{1\linewidth}
{\includegraphics[width= 1\textwidth]{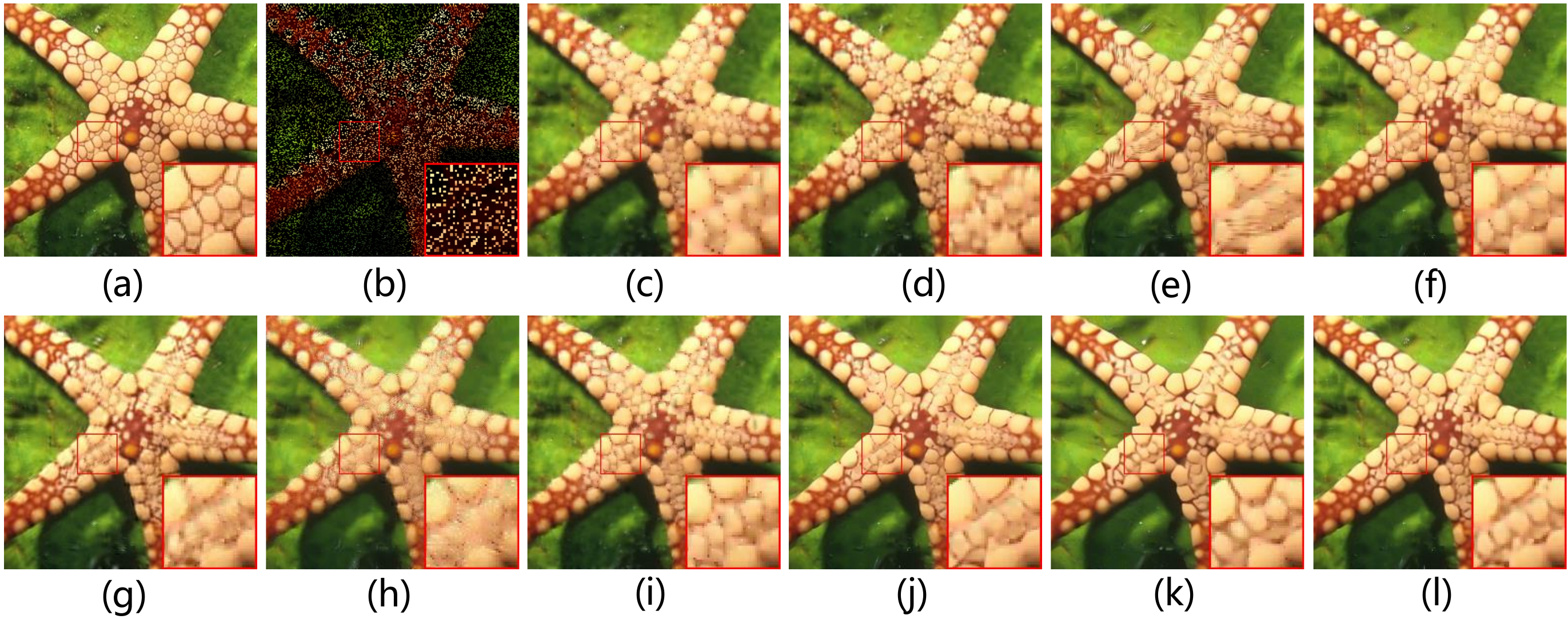}}
\end{minipage}
	\vspace{-6mm}
	\caption{Inpainting performance comparison on the image $\emph{Starfish}$. (a) Original image; (b) Degraded image with 80\% pixels missing sample; (c) SALSA \cite{41} (PSNR = 25.70dB); (d) BPFA \cite{42} (PSNR = 26.79dB); (e) IPPO \cite{43} (PSNR = 26.30dB); (f) JSM  \cite{44} (PSNR = 27.07dB); (g) Aloha \cite{45} (PSNR =26.33dB);  (h) BKSVD \cite{64} (PSNR = 25.36dB);   (i) NGS \cite{65} (PSNR = 26.17dB); (j) IRCNN \cite{70} (PSNR = 27.01dB);   (k) IDBP \cite{71} (PSNR = 26.88dB);  (l) GSC-$w\ell_p$ (PSNR = \textbf{28.05dB}).}
	\label{fig:13}
	\vspace{-4mm}
\end{figure}

{In image inpainting,  we compare the proposed GSC-$w\ell_p$  with nine state-of-the-art methods: SALSA \cite{41}, BPFA \cite{42}, IPPO \cite{43}, JSM \cite{44}, Aloha \cite{45}, BKSVD \cite{64}, NGS \cite{65}, IRCNN \cite{70} and IDBP \cite{71}. Note that BPFA, IPPO, JSM and Aloha are the image inpainting baseline methods. IRCNN and IDBP are the deep learning based methods, which use strong deep convolutional neural network (CNN). Table~\ref{lab:7} lists the PSNR results for a  collection  of 8 color images for these methods. It can be seen that the proposed GSC-$w\ell_p$  outperforms the other competing methods in most cases.  The average gains of the proposed GSC-$w\ell_p$ over SALSA, BPFA, IPPO, JSM, Aloha, BKSVD, NGS, IRCNN and IDBP methods are as much as 4.06dB, 2.26dB, 1.06dB, 1.25dB, 1.62dB, 2.93dB, 2.70dB, 0.43dB and 1.21dB, respectively. The visual comparisons of image $\emph{Zebra}$ and  image $\emph{Starfish}$  with 80\% pixels missing are shown in Fig.~\ref{fig:8} and Fig.~\ref{fig:13}, respectively. It can be seen that SALSA and BKSVD could not reconstruct sharp edges and  fine details. The BPFA, IPPO, JSM, Aloha, NGS, IRCNN and IDBP methods produce the restored images with a much better visual quality than SALSA and BKSVD, but still suffer from some undesirable artifacts, such as the ringing effects. The proposed GSC-$w\ell_p$  not only preserves sharp edges and fine details, but also eliminates the ringing effects.}

\begin{table*}[!t]
\vspace{-3mm}
	\caption{PSNR ($\textnormal{d}$B) comparison of BCS \cite{40}, RCOS \cite{46}, ASNR \cite{47}, ALSB \cite{48}, SGSR \cite{49},  JASR \cite{50}, NGS \cite{65}, PMDSE \cite{66} and GSC-$w\ell_p$ for image CS recovery.}
	\centering

	\resizebox{0.80\textwidth}{!}
	{
		\begin{tabular}{|c|c|c|c|c|c|c|c|c|c|c|c|}
			\hline
			\multirow{1}{*}{\textbf{Ratio}} &\multirow{1}{*}{\textbf{Methods}}&\multirow{1}{*}{\textbf{Barbara}}&\multirow{1}{*}{\textbf{Bridge}}
			&\multirow{1}{*}{\textbf{Elaine}}&\multirow{1}{*}{\textbf{Fence}}
			&\multirow{1}{*}{\textbf{House}}&\multirow{1}{*}{\textbf{Lena}}
			&\multirow{1}{*}{\textbf{Peppers}}&\multirow{1}{*}{\textbf{Straw}}&\multirow{1}{*}{\textbf{Average}}\\
			\cline{2-11}
			\hline
			\multirow{9}{*}{0.2}
			&   BCS          & 24.24 & 23.61 & 31.18 & 21.57 & 30.54 & 28.15 & 27.15 & 20.69 & 25.89\\
			\cline{2-11}
			& RCOS            & 27.01 & 25.09 & 33.85 & 27.31 & 35.21 & 30.33 & 30.78 & 21.72 & 28.91\\
			\cline{2-11}
			& ASNR            & 33.29 & 25.26 & 35.92 & 29.74 & \textbf{37.15} & 31.10 & 31.12 & 25.03 & 31.08\\
			\cline{2-11}
			& ALSB             & 30.72 & 24.97 & 32.56 & 28.41 & 36.08 & 30.68 & 29.96 & 24.33 & 29.71\\
			\cline{2-11}
			& SGSR              & 33.44 & 24.72 & 34.85 & 29.42 & 35.81 & 30.89 & 30.51 & 24.54 & 30.52\\
			\cline{2-11}
		
			& JASR             & 34.16 & 25.18 & 35.66 & 29.95 & 35.88 & 31.19 & 31.06 & 24.95 & 31.00\\
			\cline{2-11}

			& NGS             &  24.61 & 24.54 & 33.39 & 23.65 & 34.79 & 29.60 & 30.56 & 20.72 & 27.73\\
			\cline{2-11}

			&   PMDSE         & 24.16 & 23.06 & 29.67 & 22.62 & 29.41 & 27.24 & 25.42 & 21.57 & 25.39\\
			\cline{2-11}
			& GSC-$w\ell_p$       & \textbf{34.55} & \textbf{25.28} & \textbf{36.00} & \textbf{30.38} & {36.92} & \textbf{31.62} & \textbf{31.32} & \textbf{25.06} & \textbf{31.39}\\
 \hline
			\multirow{9}{*}{0.3}
			&   BCS          & 25.59 & 25.00 & 33.68 & 23.24 & 32.85 & 30.16 & 29.05 & 22.19 & 27.72\\
			\cline{2-11}
			& RCOS           & 30.10 & 26.42 & 36.39 & 29.91 & 37.10 & 32.38 & 32.69 & 23.96 & 31.12\\
			\cline{2-11}
			& ASNR           & 36.03 & 27.09 & 38.00 & 31.95 & 39.04 & 33.84 & 32.99 & 27.68 & 33.33\\
			\cline{2-11}
			& ALSB              & 35.00 & 26.83 & 34.30 & 30.83 & 38.34 & 33.36 & 32.37 & 26.61 & 32.20\\
			\cline{2-11}
			& SGSR            & 35.91 & 26.80 & 36.87 & 31.56 & 37.37 & 33.27 & 32.71 & 27.33 & 32.73\\
			\cline{2-11}

			& JASR              & 36.59 & 27.19 & 36.83 & 31.87 & 38.04 & 34.05 & 33.09 & 27.87 & 33.19\\
			\cline{2-11}

			& NGS             &  27.28 & 25.99 & 35.89 & 27.61 & 36.43 & 32.45 & 32.90 &  32.41 & 30.12\\
			\cline{2-11}

			&   PMDSE            & 26.73 & 24.96 & 32.31 & 25.31 & 32.05 & 29.09 & 27.57 & 24.08 & 27.76\\
			\cline{2-11}
			& GSC-$w\ell_p$       & \textbf{37.23} & \textbf{27.22} & \textbf{38.30} & \textbf{32.53} & \textbf{39.23} & \textbf{34.29} & \textbf{33.32} & \textbf{27.89} & \textbf{33.75}\\
 \hline
			\multirow{9}{*}{0.4}
			&   BCS           & 27.10 & 26.31 & 35.66 & 24.81 & 34.65 & 32.06 & 30.77 & 23.71 & 29.38\\
			\cline{2-11}
			&  RCOS           & 33.16 & 27.93 & 38.25 & 32.19 & 38.60 & 34.31 & 34.26 & 25.90 & 33.07\\
			\cline{2-11}
			& ANSR          & 38.34 & 28.76 & 39.78 & 33.92 & 40.81 & 36.09 & 34.92 & 30.04 & 35.33\\
			\cline{2-11}
			& ALSB              & 38.34 & 29.10 & 39.60 & 32.83 & 40.25 & 35.47 & 34.44 & 28.54 & 34.58\\
			\cline{2-11}
			& SGSR             & 37.70 & 28.46 & 38.63 & 33.34 & 38.99 & 35.68 & 34.47 & 29.63 & 34.61\\
			\cline{2-11}

			& JASR               & 37.39 & 28.69 & 38.28 & 33.96 & 38.80 & 36.12 & \textbf{34.70} & 30.04 & 34.75\\
			\cline{2-11}

			& NGS                &  30.43 & 27.59 & 37.76 & 30.59 & 37.82 & 35.10 & 34.57 &  24.65 & 32.31\\
			\cline{2-11}

			&  PMDSE            & 29.74 & 26.47 & 34.27 & 27.66 & 33.94 & 30.97 & 28.94 & 25.92 & 29.74\\
			\cline{2-11}
			& GSC-$w\ell_p$       & \textbf{39.13} & \textbf{28.85} & \textbf{40.05} & \textbf{34.42} & \textbf{40.93} & \textbf{36.66} & \textbf{35.00} & \textbf{30.28} & \textbf{35.67}\\
\hline
			\multirow{9}{*}{0.5}
			&   BCS         & 28.67 & 27.64 & 37.51 & 26.20 & 36.29 & 33.78 & 32.31 & 25.30 & 30.96\\
			\cline{2-11}
			& RCOS          & 35.36 & 29.03 & 39.73 & 33.82 & 40.02 & 36.19 & 35.89 & 28.02 & 34.76\\
			\cline{2-11}
			& ASNR           & 40.24 & 30.41 & 41.37 & 35.68 & 42.32 & 38.39 & 36.42 & 32.06 & 37.11\\
			\cline{2-11}
			& ALSB              & 39.26 & 30.03 & 41.18 & 34.81 & 41.93 & 37.79 & 36.21 & 30.61 & 36.48\\
			\cline{2-11}
			& SGSR             & 39.38 & 30.09 & 40.07 & 35.28 & 40.56 & 37.90 & 35.97 & 31.71 & 36.37\\
			\cline{2-11}

			& JASR               & 40.31 & 30.30 & 39.47 & 35.72 & 41.44 & 38.33 & 36.22 & 32.04 & 36.73\\
			\cline{2-11}

			& NGS                &  33.31 & 29.32 & 39.41 & 33.68 & 39.44 & 37.21 & 36.07 &  26.11 & 34.32\\
			\cline{2-11}

			&   PMDSE           & 31.64 & 27.70 & 35.79 & 29.55 & 34.79 & 32.09 & 30.09 & 26.95 & 31.07\\
			\cline{2-11}
			& GSC-$w\ell_p$       & \textbf{40.94} & \textbf{30.52} & \textbf{41.63} & \textbf{36.24} & \textbf{42.38} & \textbf{39.09} & \textbf{36.53} & \textbf{32.46} & \textbf{37.47}\\
			\hline
	\end{tabular}}
	\label{lab:8}
	\vspace{-2mm}
\end{table*}

\begin{figure}[!t]
\begin{minipage}[b]{1\linewidth}
{\includegraphics[width= 1\textwidth]{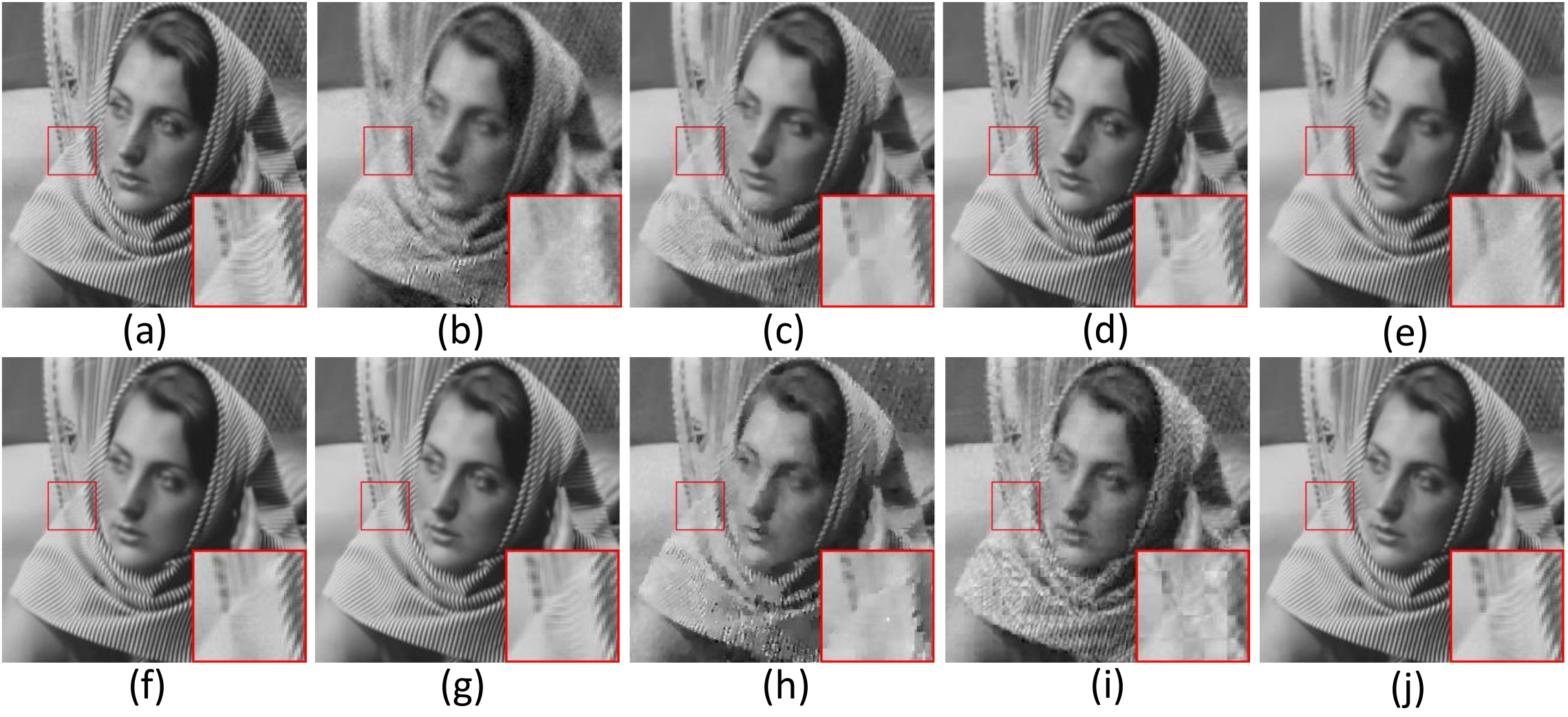}}
\end{minipage}
	\vspace{-6mm}
	\caption{CS recovery of $\emph{Barbara}$ with 0.2$N$ measurements. (a) Original image; (b) BCS \cite{40} (PSNR =24.24dB); (c) RCOS \cite{46} (PSNR= 27.01dB); (d) ASNR \cite{47}  (PSNR= 33.29dB); (e) ALSB \cite{48} (PSNR= 30.72dB); (f) SGSR \cite{49} (PSNR= 33.44dB); (g) JASR \cite{50} (PSNR= 34.16dB); (h) NGS \cite{65} (PSNR = 24.61dB);  (i) PMDSE \cite{66} (PSNR = 24.16dB);   (j) GSC-$w\ell_p$ (PSNR= \textbf{34.55dB}).}
	\label{fig:9}
	\vspace{-5mm}
\end{figure}

\begin{figure}[!t]
\begin{minipage}[b]{1\linewidth}
{\includegraphics[width= 1\textwidth]{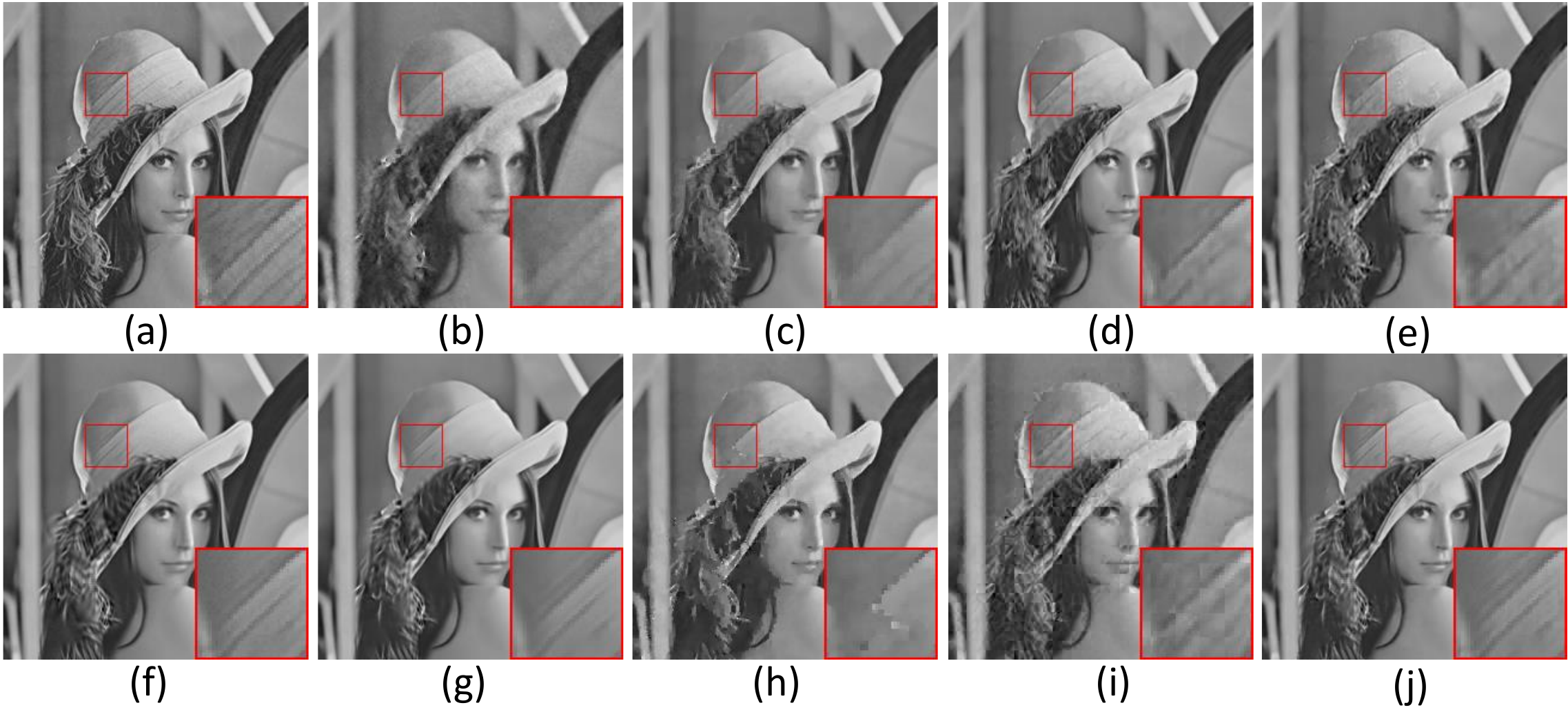}}
\end{minipage}
	\vspace{-6mm}
	\caption{CS recovery of $\emph{Lena}$ with 0.2$N$ measurements. (a) Original image; (b) BCS \cite{40} (PSNR =28.15dB); (c) RCOS \cite{46} (PSNR= 30.33dB); (d) ASNR \cite{47} (PSNR= 31.10dB); (e) ALSB \cite{48} (PSNR= 30.68dB); (f) SGSR \cite{49} (PSNR= 30.89dB); (g) JASR \cite{50} (PSNR= 31.19dB); (h) NGS \cite{65} (PSNR = 29.60dB);  (i) PMDSE \cite{66} (PSNR = 27.24dB);   (j) GSC-$w\ell_p$ (PSNR= \textbf{31.62dB}).}
	\label{fig:14}
	\vspace{-2mm}
\end{figure}

In image CS recovery, we compare the proposed GSC-$w\ell_p$ with eight competitive methods including BCS \cite{40}, RCOS \cite{46}, ASNR \cite{47}, ALSB \cite{48}, SGSR \cite{49},  JASR \cite{50}, NGS \cite{65} and PMDSE \cite{66}. Note that JASR is a recently block-based image CS recovery method that delivers state-of-the-art reconstruction results. The PSNR results are shown in Table~\ref{lab:8}. One can observe that the proposed GSC-$w\ell_p$ consistently outperforms the other competing methods in terms of PSNR (the only exception is the image $\emph{House}$ with 0.2$N$ measurements for which ASNR slightly outperforms the proposed GSC-$w\ell_p$). The proposed GSC-$w\ell_p$ achieves 6.08dB, 2.61dB, 0.36dB, 1.33dB, 1.01dB, 0.65dB, 3.45dB and 6.08dB improvements on average over  BCS, RCOS, ASNR, ALSB, SGSR, JASR,  NGS and PMDSE methods, respectively. The visual comparisons of image $\emph{Barbara}$ and image $\emph{Lena}$  with 0.2$N$ measurements are shown in Fig.~\ref{fig:9} and Fig.~\ref{fig:14}, respectively. One can observe that BCS and PMDSE methods cannot obtain the well perceptual results. Though RCOS, ASNR, ALSB, SGSR, JASR and NGS methods obtain better results than  BCS and PMDSE methods, they still suffer from some undesirable visual artifacts or over-smooth effects. The proposed GSC-$w\ell_p$  not only removes most of the visual artifacts, but also preserves large-scale sharp edges and small-scale fine image details.

\begin{table*}[!t]
\vspace{-6mm}
	\caption{{Testing the different power $p$ values for the influence of different image restoration tasks.}}
	\centering
\tiny
	\resizebox{1\textwidth}{!}
	{
	\begin{tabular}{|c|c|c|c|c|c|c|c|c|c|c|c|c|c|c|c|c|c|c|c|c|c|c|c|c|c|c|c|c|c|}
		\hline
		\multicolumn{21}{|c|}{Image Inpainting}\\
		\hline
		\multirow{1}{*}{{{p}}}&\multirow{1}{*}{{{0.05}}}&\multirow{1}{*}{{{0.1}}}&\multirow{1}{*}{{{0.15}}}
		&{{{0.2}}}&\multirow{1}{*}{{{0.25}}}&\multirow{1}{*}{{{0.3}}} &\multirow{1}{*}{{{0.35}}} &\multirow{1}{*}{{{0.4}}}
        &\multirow{1}{*}{{{0.45}}}&\multirow{1}{*}{{{0.5}}}&\multirow{1}{*}{{{0.55}}} &\multirow{1}{*}{{{0.6}}}&\multirow{1}{*}{{{0.65}}}
        &\multirow{1}{*}{{{0.7}}}&\multirow{1}{*}{{{0.75}}}&\multirow{1}{*}{{{0.8}}}&\multirow{1}{*}{{{0.85}}}&\multirow{1}{*}{{{0.9}}}
        &\multirow{1}{*}{{{0.95}}}&\multirow{1}{*}{{{1}}}\\
		\hline
\multirow{1}{*}{80\% }     &	25.69 	&	25.76 	&	25.83 	&	25.89 	&	25.94 	&	25.99 	&	26.02 	&	26.04 	&	\textbf{26.08} 	&	26.07 	&	26.06 	&	26.05 	&	26.03 	&	25.99 	&	25.92 	&	25.82 	&	25.72 	&	25.53 	&	25.38 	&	25.16 \\
 	   \hline
\multirow{1}{*}{70\% }    &	28.24 	&	28.30 	&	28.36 	&	28.41 	&	28.45 	&	28.49 	&	28.51 	&	28.54 	&	\textbf{28.55} 	&	28.55 	&	28.48 	&	28.45 	&	28.41 	&	28.35 	&	28.24 	&	28.15 	&	28.01 	&	27.89 	&	27.71 	&	27.44 \\
 	   \hline
\multirow{1}{*}{60\% }    &	28.54 	&	28.70 	&	28.84 	&	28.98 	&	29.13 	&	29.26 	&	29.38 	&	29.49 	&	29.58 	&	29.66 	&	29.75 	&	29.82 	&	29.90 	&	29.97 	&	30.04 	&	30.09 	&	30.14 	&	30.18 	&	\textbf{30.22} 	&	30.20\\
 	   \hline
\multirow{1}{*}{50\% }    &	30.53 	&	30.72 	&	30.90 	&	31.07 	&	31.21 	&	31.34 	&	31.47 	&	31.57 	&	31.67 	&	31.76 	&	31.84 	&	31.92 	&	31.99 	&	32.05 	&	32.11 	&	32.15 	&	32.18 	&	32.21 	&	\textbf{32.23} 	&	32.22\\
 	   \hline
		\multicolumn{21}{|c|}{Image CS Recovery}\\
		\hline
		\multirow{1}{*}{{{p}}}&\multirow{1}{*}{{{0.05}}}&\multirow{1}{*}{{{0.1}}}&\multirow{1}{*}{{{0.15}}}
		&{{{0.2}}}&\multirow{1}{*}{{{0.25}}}&\multirow{1}{*}{{{0.3}}} &\multirow{1}{*}{{{0.35}}} &\multirow{1}{*}{{{0.4}}}
        &\multirow{1}{*}{{{0.45}}}&\multirow{1}{*}{{{0.5}}}&\multirow{1}{*}{{{0.55}}} &\multirow{1}{*}{{{0.6}}}&\multirow{1}{*}{{{0.65}}}
        &\multirow{1}{*}{{{0.7}}}&\multirow{1}{*}{{{0.75}}}&\multirow{1}{*}{{{0.8}}}&\multirow{1}{*}{{{0.85}}}&\multirow{1}{*}{{{0.9}}}
        &\multirow{1}{*}{{{0.95}}}&\multirow{1}{*}{{{1}}}\\
		\hline
\multirow{1}{*}{0.2$N$}&	28.40 	&	28.48 	&	28.57 	&	28.64 	&	28.72 	&	28.79 	&	28.86 	&	28.91 	&	28.96 	&	\textbf{29.06} 	&	29.04 	&	29.01 	&	28.99 	&	28.98 	&	28.89 	&	28.62 	&	28.50 	&	28.14 	&	27.86 	&	27.67\\
 	   \hline
\multirow{1}{*}{0.3$N$}&	30.40 	&	30.47 	&	30.55 	&	30.63 	&	30.71 	&	30.79 	&	30.88 	&	30.96 	&	31.05 	&	31.14 	&	31.23 	&	31.32 	&	31.41 	&	31.51 	&	31.60 	&	31.68 	&	31.77 	&	31.84 	&	\textbf{31.95} 	&	31.90 \\
\hline
\multirow{1}{*}{0.4$N$}&	32.93 	&	33.03 	&	33.12 	&	33.21 	&	33.30 	&	33.38 	&	33.47 	&	33.55 	&	33.64 	&	33.72 	&	33.80 	&	33.87 	&	33.94 	&	34.00 	&	34.05 	&	34.10 	&	34.14 	&	34.17 	&	\textbf{34.19} 	&	34.18\\
\hline
\multirow{1}{*}{0.5$N$}&	34.91 	&	35.01 	&	35.10 	&	35.20 	&	35.30 	&	35.40 	&	35.50 	&	35.59 	&	35.69 	&	35.78 	&	35.86 	&	35.93 	&	35.99 	&	36.05 	&	36.10 	&	36.14 	&	36.16 	&	36.16 	&	\textbf{36.17} 	&	36.16\\
\hline
	\end{tabular}}
\vspace{-2mm}
	\label{lab:9}
\end{table*}

\subsection{Suitable setting of the power $p$}
\label{5.4}
{In this subsection, we discuss how to select $p$ value for the superior performance of the proposed GSC-$w\ell_p$. To be concrete, we randomly select 30 images (size: 256$\times$ 256) and test the proposed GSC-$w\ell_p$ with different $p$ values under the image inpainting and image CS recovery tasks. Table~\ref{lab:9} shows the average PSNR results under different $p$ values from 0.05 to 1 with interval 0.05. It can be seen that the best reconstruction performance of the image inpainting task is obtained by setting $p = 0.45, 0.45, 0.95$ and 0.95 when 80\%, 70\%, 60\% and 50\% pixels are missing, respectively. Meanwhile, the best CS reconstruction performance of the image CS recovery task is obtained by setting $p = 0.5, 0.95, 0.95$ and 0.95 with 0.2$N$, 0.3$N$, 0.4$N$ and 0.5$N$ measurements, respectively. Therefore, in image inpainting, we choose $p = 0.45, 0.45, 0.95$ and 0.95 when 80\%, 70\%, 60\% and 50\% pixels are missing, respectively. In image CS recovery, we choose $p = 0.5, 0.95, 0.95$ and 0.95 with 0.2$N$, 0.3$N$, 0.4$N$ and 0.5$N$ measurements, respectively.}

\subsection{Convergence}
\begin{figure}[!t]
	\centering
\begin{minipage}[b]{1\linewidth}
{\includegraphics[width= 1\textwidth]{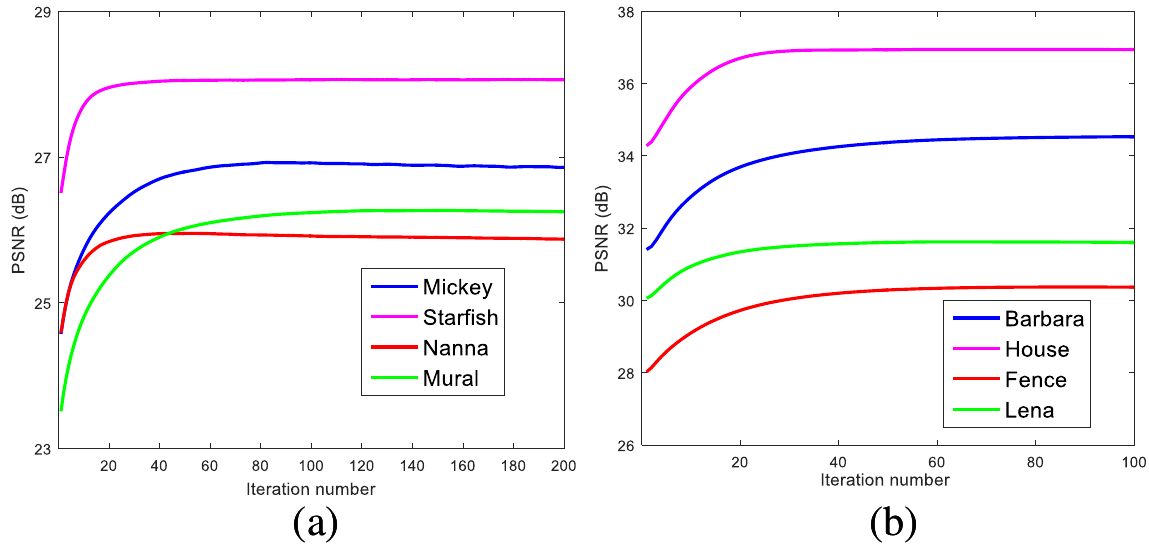}}
\end{minipage}
\vspace{-6mm}
	\caption{Convergence analysis of the proposed scheme. (a) PSNR results versus iteration number for image inpainting with 80\%  pixels missing. (b) PSNR results versus iteration number for image CS recovery with 0.2$N$ measurements. }
	\vspace{-2mm}
	\label{fig:12}
\end{figure}

Since the proposed model is non-convex, it is difficult to provide its theoretical proof of global convergence. Hereby, we present empirical evidence to show the convergence of the proposed model. Fig.~\ref{fig:12} plots the curves of the PSNR values versus the iteration numbers for image inpainting with 80\% pixels missing for image $\emph{Mickey}, \emph{Starfish}, \emph{Nanna}$ and $\emph{Murla}$ as well as image CS recovery (including image $\emph{Barbara}, \emph{House}, \emph{Fence}$ and $\emph{Lena}$) with 0.2$N$ measurements, respectively. It can be seen that with the increase of the iteration numbers, the PSNR curves of the reconstructed images gradually increase and then become flat and stable. Therefore, we conclude that the proposed GSC-$w\ell_p$ has a good convergence performance.

\section{Conclusion}
\label{sec:6}
This paper proposed a benchmark for sparse coding from the perspective of rank minimization. A group-based adaptive dictionary learning method has been designed to bridge the gap between the GSC and rank minimization models. We have proved the equivalence of the GSC and rank minimization models under the designed dictionary, and thus the sparse coefficients of each patch group were measured by computing the singular values of each patch group. In this way, we have earned a benchmark to measure the sparsity of each patch group since the singular values of the original image patch groups can be easily computed by SVD operator. The proposed benchmark can be used to measure different norm minimization methods in sparse coding through analyzing their corresponding rank minimization methods. To be concrete, four well-known rank minimization methods including NNM, SNM, WNNM and WSNM, have been adopted to analyze the sparsity of each patch group and the solution of WSNM has been found to be the best approximation to real singular values of each patch group. Based on the above analysis, NNM, SNM, WNNM and WSNM can be equivalently transformed into $\ell_1$-norm minimization, $\ell_p$-norm minimization, weighted $\ell_1$-norm minimization and the weighted $\ell_p$-norm minimization problems in GSC, respectively. Meanwhile, based on the earned benchmark in sparse coding, the weighted $\ell_p$-norm minimization was also expected to achieve better performance than the  three other  norm minimization methods. To verify the feasibility of the proposed benchmark, we compared the weighted $\ell_p$-norm minimization with the three other  norm minimization methods in sparse coding. To make the proposed optimization problem tractable, we have employed the ADMM algorithm to solve the non-convex weighted $\ell_p$-norm minimization problem. Experimental results on two applications: image inpainting and image CS recovery, have demonstrated that the proposed scheme is feasible and achieves performance improvements over the state-of-the-art methods both quantitatively and qualitatively.


\appendices
\section{Proof of the Lemma~\ref{lemma:3}}
\label{lemma3}
\begin{proof}
The adaptive dictionary ${\textbf{\emph{D}}}_i$ is constructed by Eq.~\eqref{eq:11}. From the unitary property of ${\textbf{\emph{U}}}_i$ and ${\textbf{\emph{V}}}_i$, we have
\begin{equation}
\begin{aligned}
&\left\|{\textbf{\emph{Y}}}_i-{\textbf{\emph{X}}}_i\right\|_F^2 = \left\|{\textbf{\emph{D}}}_i({\textbf{\emph{B}}}_i-{\textbf{\emph{A}}}_i)\right\|_F^2
=\left\|{\textbf{\emph{U}}}_i{\rm diag}({\textbf{\emph{B}}}_i-{\textbf{\emph{A}}}_i){\textbf{\emph{V}}}_i^T\right\|_F^2\\
&= {\rm Tr}({\textbf{\emph{U}}}_i{\rm diag}({\textbf{\emph{B}}}_i-{\textbf{\emph{A}}}_i){\textbf{\emph{V}}}_i^T
{\textbf{\emph{V}}}_{i}{\rm diag}({\textbf{\emph{B}}}_i-{\textbf{\emph{A}}}_i){\textbf{\emph{U}}}_{i}^T)\\
&= {\rm Tr}({\textbf{\emph{U}}}_i{\rm diag}({\textbf{\emph{B}}}_i-{\textbf{\emph{A}}}_i)
{\rm diag}({\textbf{\emph{B}}}_i-{\textbf{\emph{A}}}_i){\textbf{\emph{U}}}_{i}^T)\\
&={\rm Tr}({\rm diag}({\textbf{\emph{B}}}_i-{\textbf{\emph{A}}}_i)
{\textbf{\emph{U}}}_i^T{\textbf{\emph{U}}}_{i}{\rm diag}({\textbf{\emph{B}}}_i-{\textbf{\emph{A}}}_i))\\
&={\rm Tr}({\rm diag}({\textbf{\emph{B}}}_i-{\textbf{\emph{A}}}_i)
{\rm diag}({\textbf{\emph{B}}}_i-{\textbf{\emph{A}}}_i))\\
&=\left\|{\textbf{\emph{B}}}_i-{\textbf{\emph{A}}}_i\right\|_F^2,
\end{aligned}
\label{eq:42}
\end{equation} 
where ${\textbf{\emph{X}}}_i={{\textbf{\emph{D}}}_i{\textbf{\emph{A}}}_i}$ and ${\textbf{\emph{Y}}}_i={{\textbf{\emph{D}}}_i{\textbf{\emph{B}}}_i}$.
\end{proof}

\section{Proof of the Theorem~\ref{theorem:1}}
\label{theorem1}
\begin{proof}
On the basis of Lemma~\ref{lemma:3}, we have
  \begin{equation}
  \begin{aligned}
{\hat{\textbf{\emph{A}}}}_i&=\arg\min\limits_{{\textbf{\emph{A}}}_i}
\left(\frac{1}{2}\left\|{\textbf{\emph{Y}}}_i-{\textbf{\emph{D}}}_i{\textbf{\emph{A}}}_i\right\|_F^2+\lambda\left\|{\textbf{\emph{A}}}_i\right\|_1\right)\\
&=\arg\min\limits_{{\textbf{\emph{A}}}_i} \left(\frac{1}{2}\left\|{\textbf{\emph{B}}}_i-{\textbf{\emph{A}}}_i\right\|_F^2+\lambda\left\|{\textbf{\emph{A}}}_i\right\|_1\right)\\
&=\arg\min\limits_{\boldsymbol\alpha_i} \left(\frac{1}{2}\left\|{\boldsymbol\beta}_i-{\boldsymbol\alpha}_i\right\|_2^2+\lambda\left\|{\boldsymbol\alpha}_i\right\|_1\right),
\end{aligned}
\label{eq:43}
\end{equation} 
where ${\textbf{\emph{X}}}_i={{\textbf{\emph{D}}}_i{\textbf{\emph{A}}}_i}$ and ${\textbf{\emph{Y}}}_i={{\textbf{\emph{D}}}_i{\textbf{\emph{B}}}_i}$. ${{{{\boldsymbol\alpha}}}_i}$ and ${{{{\boldsymbol\beta}}}_i}$ denote the vectorization of the matrix ${{{\textbf{\emph{A}}}}_i}$ and ${{{\textbf{\emph{B}}}}_i}$, respectively.

Therefore, based on Lemma~\ref{lemma:1}, we have
\begin{equation}
\begin{aligned}
{{\boldsymbol\alpha}}_i & ={\rm soft}({{\boldsymbol\beta}}_i,\lambda)= {\rm sgn}({{\boldsymbol\beta}}_i)\odot{\rm max}({\rm abs}({{\boldsymbol\beta}}_i)-\lambda,0).
\end{aligned}
\label{eq:44}
\end{equation}
Obviously, according to Eqs.~\eqref{eq:10} and ~\eqref{eq:11}, we have
\begin{equation}
\begin{aligned}
{\textbf{\emph{D}}}_i{\hat{\textbf{\emph{A}}}}_i&= \sum\nolimits_{j=1}^{n_1} {{\rm soft}({\boldsymbol\beta}}_{i,j},\lambda){\textbf{\emph{d}}}_{i,j}\\
&=\sum\nolimits_{j=1}^{n_1} {{\rm soft}({\boldsymbol\beta}}_{i,j},\lambda){\textbf{\emph{u}}}_{i,j}{\textbf{\emph{v}}}_{i,j}^T \\
&={\textbf{\emph{U}}}_i\mathcal{D}_{\lambda}(\boldsymbol\Delta_i){\textbf{\emph{V}}}_i^T =  \mathcal{D}_{\lambda}({\textbf{\emph{Y}}}_i),
\end{aligned}
\label{eq:45}
\end{equation} 
where ${{\boldsymbol\beta}}_{i,j}$ represents the $j$-th element of the $i$-th group sparse coefficient ${{\boldsymbol\beta}}_{i}$, and $\boldsymbol\Delta_i$ is the  singular value matrix of the $i$-th group ${\textbf{\emph{Y}}}_i$.

Following this, and based on Theorem~\ref{lemma:2}, we have proved that the $\ell_1$-norm minimization based GSC problem is equivalent to the NNM problem, \ie,
  \begin{equation}
  \begin{aligned}
{\hat{\textbf{\emph{A}}}}_i&=\arg\min\limits_{{\textbf{\emph{A}}}_i}
\left(\frac{1}{2}\left\|{\textbf{\emph{Y}}}_i-{\textbf{\emph{D}}}_i{\textbf{\emph{A}}}_i\right\|_F^2+\lambda\left\|{\textbf{\emph{A}}}_i\right\|_1\right)\\
&\ \ \ \ \ \ \ \ \ \ \ \ \ \ \ \ \ \ \ \ \ \ \ \ \ \ \ \ \ \Updownarrow\\
{\hat{\textbf{\emph{X}}}}_i&= \arg\min_{\textbf{\emph{X}}_i} \left(\frac{1}{2}\left\|{\textbf{\emph{Y}}}_i-{\textbf{\emph{X}}}_i\right\|_F^2+\lambda\left\|{\textbf{\emph{X}}}_i\right\|_*\right).
\end{aligned}
\label{eq:46}
\end{equation} 
\end{proof}

\section{Proof of the Corollary~\ref{coeollary:1}}
\label{coeolllar}
\begin{proof}
We first prove the weighted $\ell_1$-norm minimization based GSC problem is equivalent to the weighted nuclear norm minimization (WNNM) problem. Specifically,  the weighted norm $||\textbf{\emph{X}}||_{{\textbf{{w}}},*}$   is used to regularize $\textbf{\emph{X}}$ and Eq.~\eqref{eq:3} can be rewritten as
\begin{equation}
\hat{\textbf{\emph{X}}}= \arg\min_{\textbf{\emph{X}}} \left(\frac{1}{2}\left\|{\textbf{\emph{Y}}} -{\textbf{\emph{X}}}\right\|_F^2+\left\|\textbf{\emph{X}}\right\|_{\rm {\textbf{w}},*}\right).
\label{eq:47}
\end{equation} 
In order to prove that the equivalence of the weighted $\ell_1$-norm minimization based GSC problem and the WNNM problem, we firstly give the following lemma and theorems.

\begin{lemma}
\label{lemma:4}
For the following optimization problem
\begin{equation}
\min_{{{\emph{x}}}_i} \sum\nolimits_{i=1}^n \left(\frac{1}{2}({{\emph{a}}}_i- {{\emph{x}}}_i)^2 + \emph{w}_i \emph{x}_i\right),
\label{eq:48}
\end{equation} 
then the global optimum of Eq.~\eqref{eq:48} \emph{is} $\hat{\emph{x}}_i= {\rm soft}(\emph{a}_i,\emph{w}_i)={\rm {max}} (\emph{a}_i-\emph{w}_i, 0)$.
\end{lemma}
\begin{proof}
See \cite{60}.
\end{proof}

\begin{theorem}
\label{theorem:3}

If the singular values $\delta_1\geq ...\geq\delta_{n_1}$ and the weights satisfy $0\leq{{\emph{w}}}_{1} \leq...{{\emph{w}}}_{n_1}$, $n_1={\rm min}{(d, m)}$, the WNNM problem in Eq.~\eqref{eq:47} has a globally optimal solution,
\begin{equation}
\hat{\textbf{\emph{X}}}=\textbf{\emph{U}}\mathcal{D}_{\textbf{\emph{w}}}({\boldsymbol\Delta}){\textbf{\emph{V}}}^T,
\label{eq:49}
\end{equation} 
where $\textbf{\emph{Y}}=\textbf{\emph{U}}{\boldsymbol\Delta}{\textbf{\emph{V}}}^T$ is the SVD of $\textbf{\emph{Y}}$ and $\mathcal{D}_{\textbf{\emph{w}}}({\boldsymbol\Delta})$ is the generalized soft-thresholding operator with the weighted vector ${\textbf{\emph{w}}}$, i.e., $\mathcal{D}_{\textbf{\emph{w}}}{({\boldsymbol\Delta})}_{i}={\rm soft}({\boldsymbol\Delta}_{{i}},\emph{w}_i)={\rm
max}({\boldsymbol\Delta}_{{i}}-\emph{w}_i,0)$, where ${\boldsymbol\Delta}_{{i}}$ denotes the $i$-th singular value of ${\boldsymbol\Delta}$.
\end{theorem}
\begin{proof}
See \cite{26}.
\end{proof}

\begin{theorem}
\label{theorem:4}
For $0\leq{{\emph{w}}}_{1}\leq...{{\emph{w}}}_{n_1}$ and $\textbf{\emph{Y}}\in\mathbb{R}^{d \times m}$, $n_1={\rm min}{(d, m)}$, the singular value shrinkage operator Eq.~\eqref{eq:49} satisfies Eq.~\eqref{eq:47}.
\end{theorem}
\begin{proof}
See \cite{54}.
\end{proof}
Then, for patch group ${\textbf{\emph{X}}}_i$ in ${\textbf{\emph{X}}}$, the WNNM problem in Eq.~\eqref{eq:47} can be rewritten as

\begin{equation}
\hat{\textbf{\emph{X}}}_i= \arg\min_{\textbf{\emph{X}}_i} \left(\frac{1}{2}\left\|{\textbf{\emph{Y}}}_i -{\textbf{\emph{X}}}_i\right\|_F^2+\left\|\textbf{\emph{X}}\right\|_{\rm {\textbf{w}}_i,*}\right).
\label{eq:47.1}
\end{equation} 

For the weighted $\ell_1$-norm minimization based GSC problem and based on Lemma~\ref{lemma:3}, we have,
\begin{equation}
\begin{aligned}
{\hat{\textbf{\emph{A}}}}_i&=\arg\min\limits_{{\textbf{\emph{A}}}_i}
\left(\frac{1}{2}\left\|{\textbf{\emph{Y}}}_i-{\textbf{\emph{D}}}_i{\textbf{\emph{A}}}_i\right\|_F^2 + \left\|{\textbf{{W}}}_i\circ{\textbf{\emph{A}}}_i\right\|_1\right)\\
&=\arg\min\limits_{{\textbf{\emph{A}}}_i} \left(\frac{1}{2}\left\|{\textbf{\emph{B}}}_i-{\textbf{\emph{A}}}_i\right\|_F^2 + \left\|{\textbf{{W}}}_i\circ {\textbf{\emph{A}}}_i\right\|_1\right)\\
&=\arg\min\limits_{\boldsymbol\alpha_i} \left(\frac{1}{2}\left\|{\boldsymbol\beta}_i-{\boldsymbol\alpha}_i\right\|_2^2+ \left\|\textbf{\emph{w}}_i{\boldsymbol\alpha}_i\right\|_1\right),
\end{aligned}
\label{eq:50}
\end{equation} 
where $\circ$ represents the element-wise product of two matrices (Hadamard product). ${\textbf{\emph{X}}}_i={{\textbf{\emph{D}}}_i{\textbf{\emph{A}}}_i}$ and ${\textbf{\emph{Y}}}_i={{\textbf{\emph{D}}}_i{\textbf{\emph{B}}}_i}$. ${{{{\boldsymbol\alpha}}}_i}$, ${{{{\boldsymbol\beta}}}_i}$ and $\textbf{\emph{w}}_i$ denote the vectorization of the matrix ${{{\textbf{\emph{A}}}}_i}$, ${{{\textbf{\emph{B}}}}_i}$ and $\textbf{{W}}_i$, respectively.

Then, based on Lemma~\ref{lemma:4}, we have,
\begin{equation}
{{\boldsymbol\alpha}}_i  ={\rm soft}({{\boldsymbol\beta}}_i,\textbf{\emph{w}}_i)=  {\rm max}({{\boldsymbol\beta}}_i-\textbf{\emph{w}}_i,0).
\label{eq:51}
\end{equation}

Based on  Eqs.~\eqref{eq:10} and ~\eqref{eq:11}, we have
\begin{equation}
\begin{aligned}
&\hat{\textbf{\emph{X}}}_i={{\textbf{\emph{D}}}_i{\textbf{\emph{A}}}_i}= \sum\nolimits_{j=1}^{n_1} {{\rm soft}({\boldsymbol\beta}}_{i,j},{{{w}}}_{i,j}){\textbf{\emph{d}}}_{i,j}\\
&=\sum\nolimits_{j=1}^{n_1} {{\rm soft}({\boldsymbol\beta}}_{i,j},{{{w}}}_{i,j}){\textbf{\emph{u}}}_{i,j}{\textbf{\emph{v}}}_{i,j}^T\\ &={\textbf{\emph{U}}}_i\mathcal{D}_{\textbf{{w}}_i}(\boldsymbol\Delta_i){\textbf{\emph{V}}}_i^T.
\end{aligned}
\label{eq:52}
\end{equation} 

Therefore, based on Theorem~\ref{theorem:4}, we prove that the weighted $\ell_1$-norm minimization based GSC problem is equivalent to the WNNM problem, \ie,
  \begin{equation}
  \begin{aligned}
{\hat{\textbf{\emph{A}}}}_i&=\arg\min\limits_{{\textbf{\emph{A}}}_i}
\left(\frac{1}{2}\left\|{\textbf{\emph{Y}}}_i-{\textbf{\emph{D}}}_i{\textbf{\emph{A}}}_i\right\|_F^2+ \left\|{\textbf{{W}}}_i\circ{\textbf{\emph{A}}}_i\right\|_1\right)\\
&\ \ \ \ \ \ \ \ \ \ \ \ \ \ \ \ \ \ \ \ \ \ \ \ \ \ \ \ \  \Updownarrow\\
{\hat{\textbf{\emph{X}}}}_i&= \arg\min_{\textbf{\emph{X}}_i} \left(\frac{1}{2}\left\|{\textbf{\emph{Y}}}_i-{\textbf{\emph{X}}}_i\right\|_F^2+ \left\|\textbf{\emph{X}}_i\right\|_{\rm {\textbf{w}_i},*}\right).
\end{aligned}
\label{eq:53}
\end{equation} 

Note that here we assume the setting of the weight $\textbf{W}_i$ in Eq.~\eqref{eq:47.1}  is as the same as the weight $\textbf{W}_i$ in Eq.~\eqref{eq:50}.

Next, we prove the $\ell_p$-norm minimization based GSC problem is equivalent to the Schatten $p$-norm minimization (SNM) problem. Specifically,  the Schatten $p$-norm $||\textbf{\emph{X}}_i||_{S_p}$   is used to regularize $\textbf{\emph{X}}_i$ in Eq.~\eqref{eq:3}  and  for patch group $\textbf{\emph{X}}_i$, we have,
\begin{equation}
\hat{\textbf{\emph{X}}}_i =\min\limits_{\textbf{\emph{X}}_i} \left(\frac{1}{2}\left\|{\textbf{\emph{Y}}}_i-{\textbf{\emph{X}}}_i\right\|_F^2 + \lambda
 \left\|\textbf{\emph{X}}_i\right\|_{S_p}\right).
\label{eq:54}
\end{equation} 

To prove that the equivalence of the $\ell_p$-norm minimization based GSC problem and the SNM problem, we firstly give the following theorem.

\begin{theorem}
\label{theorem:5}
Let $\textbf{\emph{Y}}_i= \textbf{\emph{U}}_i\boldsymbol\Delta_i\textbf{\emph{V}}_i^T$ be the SVD of $\textbf{\emph{Y}}_i\in\mathbb{R}^{d\times m}$ and $\boldsymbol\Delta_i=diag(\delta_{i,1}, ..., \delta_{i,n_1})$, $n_1= min(d,m)$. The optimal solution $\textbf{\emph{X}}_i$ to problem Eq.~\eqref{eq:54} is $\textbf{\emph{U}}_i\boldsymbol\Sigma_i\textbf{\emph{V}}_{i}^T$, where $\boldsymbol\Sigma_{i}=diag(\sigma_{i,1}, ..., \sigma_{i,n_1})$. Then the solution of the $j$-th diagonal element $\sigma_{i,j}$ of the diagonal matrix $\boldsymbol\Sigma_{i}$ is solved by the following problem,
\begin{equation}
\begin{aligned}
&\min\limits_{\sigma_{i,j}}
\left(\frac{1}{2}(\delta_{i,j}-\sigma_{i,j})^2+\lambda \sigma_{i,j}^p \right),
\end{aligned}
\label{eq:55}
\end{equation}
where $\sigma_{i,j}$ represents the $j$-th singular value of each data matrix $\textbf{\emph{X}}_i$.
\end{theorem}

\begin{proof}
See \cite{25}.
\end{proof}

Now, for the $\ell_p$-norm minimization based GSC problem, and based on Lemma~\ref{lemma:3}, we have,
\begin{equation}
\begin{aligned}
{\hat{\textbf{\emph{A}}}}_i&=\arg\min\limits_{{\textbf{\emph{A}}}_i}
\left(\frac{1}{2}\left\|{\textbf{\emph{Y}}}_i-{\textbf{\emph{D}}}_i{\textbf{\emph{A}}}_i\right\|_F^2 + \lambda \left\|{\textbf{\emph{A}}}_i\right\|_p\right)\\
&=\arg\min\limits_{{\textbf{\emph{A}}}_i} \left(\frac{1}{2}\left\|{\textbf{\emph{B}}}_i-{\textbf{\emph{A}}}_i\right\|_F^2 + \lambda\left\| {\textbf{\emph{A}}}_i\right\|_p\right)\\
&=\arg\min\limits_{\boldsymbol\alpha_i} \left(\frac{1}{2}\left\|{\boldsymbol\beta}_i- {\boldsymbol\alpha}_i\right\|_2^2+ \lambda\left\|{\boldsymbol\alpha}_i\right\|_p\right),
\end{aligned}
\label{eq:56}
\end{equation} 
where ${\textbf{\emph{X}}}_i={{\textbf{\emph{D}}}_i{\textbf{\emph{A}}}_i}$ and ${\textbf{\emph{Y}}}_i={{\textbf{\emph{D}}}_i{\textbf{\emph{B}}}_i}$. ${{{{\boldsymbol\alpha}}}_i}$ and ${{{{\boldsymbol\beta}}}_i}$ denote the vectorization of the matrix ${{{\textbf{\emph{A}}}}_i}$ and ${{{\textbf{\emph{B}}}}_i}$, respectively.

Then the solution of the $j$-th element $\boldsymbol\alpha_{i,j}$ of  $\boldsymbol\alpha_i$ in Eq.~\eqref{eq:56} is solved by the following minimization problem,
\begin{equation}
\begin{aligned}
&\min\limits_{\boldsymbol\alpha_{i,j}}
\left(\frac{1}{2}(\boldsymbol\beta_{i,j}-\boldsymbol\alpha_{i,j})^2+\lambda \boldsymbol\alpha_{i,j}^p \right).
\end{aligned}
\label{eq:57.1}
\end{equation}

Obviously, based on the designed adaptive dictionary $\textbf{\emph{D}}_i$ in Eq.~\eqref{eq:11}, Eq.~\eqref{eq:57.1} is equivalent to Eq.~\eqref{eq:55}.

Therefore, we prove that the $\ell_p$-norm minimization based GSC problem is equivalent to the SNM problem, \ie,
\begin{equation}
\begin{aligned}
{\hat{\textbf{\emph{A}}}}_i&=\arg\min\limits_{{\textbf{\emph{A}}}_i}
\left(\frac{1}{2}\left\|{\textbf{\emph{Y}}}_i-{\textbf{\emph{D}}}_i{\textbf{\emph{A}}}_i\right\|_F^2+ \lambda \left\|{\textbf{\emph{A}}}_i\right\|_p\right)\\
&\ \ \ \ \ \ \ \ \ \ \ \ \ \ \ \ \ \ \ \ \ \ \ \ \ \ \ \ \ \Updownarrow\\
{\hat{\textbf{\emph{X}}}}_i&= \arg\min_{\textbf{\emph{X}}_i} \left(\frac{1}{2}\left\|{\textbf{\emph{Y}}}_i-{\textbf{\emph{X}}}_i\right\|_F^2+ \lambda \left\|\textbf{\emph{X}}_i\right\|_{S_p}\right).
\end{aligned}
\label{eq:58}
\end{equation} 

Similar to the $\ell_p$-norm minimization based GSC problem is equivalent to the SNM problem, we can also prove that the weighted $\ell_p$-norm minimization based GSC problem is equivalent to the weighted Schatten $p$-norm minimization (WSNM) problem, \ie,
\begin{equation}
\begin{aligned}
{\hat{\textbf{\emph{A}}}}_i&=\arg\min\limits_{{\textbf{\emph{A}}}_i}
\left(\frac{1}{2}\left\|{\textbf{\emph{Y}}}_i-{\textbf{\emph{D}}}_i{\textbf{\emph{A}}}_i\right\|_F^2+  \left\|\textbf{W}_i\circ{\textbf{\emph{A}}}_i\right\|_p\right)\\
&\ \ \ \ \ \ \ \ \ \ \  \ \ \ \ \ \ \ \ \ \ \ \  \ \ \ \ \ \  \Updownarrow\\
{\hat{\textbf{\emph{X}}}}_i&= \arg\min_{\textbf{\emph{X}}_i} \left(\frac{1}{2}\left\|{\textbf{\emph{Y}}}_i-{\textbf{\emph{X}}}_i\right\|_F^2+  \left\|\textbf{\emph{X}}_i\right\|_{\textbf{{w}}_i,S_p}\right).
\end{aligned}
\label{eq:59}
\end{equation} 
Therefore, based on the above analysis, we prove the Corollary~\ref{coeollary:1}.
\end{proof}

\section*{Acknowledgment}
The authors would like to appreciate the associate editor for coordinating the review of the manuscript, and appreciate the anonymous reviewers for their constructive suggestions to improve the manuscript. The authors would like to appreciate Prof. Jian Zhang at Peking University for his help.

{\footnotesize
\bibliographystyle{IEEEtran}
\bibliography{gwsnm_ref}
}

\end{document}